\documentclass{article}

\usepackage{PRIMEarxiv}

\usepackage[utf8]{inputenc} % allow utf-8 input
\usepackage[T1]{fontenc}    % use 8-bit T1 fonts
\usepackage{hyperref}       % hyperlinks
\usepackage{url}            % simple URL typesetting
\usepackage{booktabs}       % professional-quality tables
\usepackage{amsfonts}       % blackboard math symbols
\usepackage{amsmath}
\usepackage{nicefrac}       % compact symbols for 1/2, etc.
\usepackage{microtype}      % microtypography
\usepackage{lipsum}
\usepackage{fancyhdr}       % header
\usepackage{graphicx}       % graphics
\usepackage{xcolor}
\usepackage{subcaption}
\usepackage{wrapfig}
\usepackage[subtle]{savetrees}
\usepackage{natbib}

\graphicspath{{media/}}     % organize your images and other figures under media/ folder

\newcommand{\R}{\mathbb{R}}

\DeclareMathOperator*{\softmax}{softmax}
  
\title{Universal Neurons in GPT2 Language Models
}

\author{
Wes Gurnee$^{1}$\thanks{Correspondence: \texttt{wesg@mit.edu}; $^\dagger$\text{Senior Author}} \quad Theo Horsley$^2$ \quad Zifan Carl Guo$^{1}$  \quad Tara Rezaei Kheirkhah$^{1}$  \\ 
\vspace{0.2em}
\textbf{Qinyi Sun}$^{1}$ \quad \textbf{Will Hathaway}$^{1}$ \quad  \textbf{Neel Nanda}$^{\dagger}$ \quad  \textbf{Dimitris Bertsimas}$^{1\dagger}$ \\
$^1$MIT \quad $^2$University of Cambridge 
}
\begin{document}
\maketitle

\begin{abstract}
% With increasing interest in interpreting the inner mechanisms of neural networks, a key question is the degree to which these models learn the same underlying mechanisms.
A basic question within the emerging field of mechanistic interpretability is the degree to which neural networks learn the same underlying mechanisms. In other words, are neural mechanisms universal across different models? 
% In other words, do different models form the same structures connected in similar ways to execute similar algorithms?
In this work, we study the universality of individual neurons across GPT2 models trained from different initial random seeds, motivated by the hypothesis that universal neurons are likely to be interpretable. In particular, we compute pairwise correlations of neuron activations over 100 million tokens for every neuron pair across five different seeds and find that 1-5\% of neurons are universal, that is, pairs of neurons which consistently activate on the same inputs. We then study these universal neurons in detail, finding that they usually have clear interpretations and taxonomize them into a small number of neuron families. We conclude by studying patterns in neuron weights to establish several universal functional roles of neurons in simple circuits: deactivating attention heads, changing the entropy of the next token distribution, and predicting the next token to (not) be within a particular set.

% End: partial taxonomification
% Within the emerging discipline of mechanistic interpretability, an 
% Something about universality being an important assumption or hypothesis within LLM interpretabily
% - Understand the degree of universality within language models neurons
% - Use this to identify fundamental motifs
% - Findings: not much universality
% - Also analyze in weight space
\end{abstract}

% keywords can be removed

% \section*{Summary of Main Results}
% \paragraph{Feature Families}
% \begin{itemize}
%     \item Neurons with high excess correlation (ie, that are universal) tend to be sparsely activating, bimodal, and therefore monosemantic
%     \item Most high of these high excess correlation neurons activate on a set of related tokens
% \end{itemize}

% \paragraph{Functional Roles}
% \begin{itemize}
%     \item There exist neurons with extreme kurtosis or variance in composition with the output vocabulary. These outlier tokens are often extremely interpretable.
%     \item Positive prediction neurons (those with positive skew) cluster in the mid-late layers and negative prediction (or suppression) neurons, primarily occur in the last layers.
%     \item Signal neurons
%     \item Scale neurons
% \end{itemize}

% \paragraph{Mysteries/other interesting observations}
% \begin{itemize}
%     \item BOS neurons don't seem to do anything except for at the very end of the network despite having very large norm
%     \item Neurons with positive cosine in/out cosine sim are way sparser than those with negative cosine sim
%     \item In general lots of duplication
% \end{itemize}

\section{Introduction}
As large language models (LLMs) become more widely deployed in high-stakes settings, our lack of understanding of why or how models make decisions creates many potential vulnerabilities and risks \citep{bommasani2021opportunities,hendrycks2023overview,bengio2023managing}.
While some claim deep learning based systems are fundamentally inscrutable, artificial neural networks seem unusually amenable to empirical science compared to other complex systems: they are fully observable, (mostly) deterministic, created by processes we control, admit complete mathematical descriptions of their form and function, can be run on any input with arbitrary modifications made to their internals, all at low cost and on computational timescales \citep{olah2021interpretability}.
An advanced science of interpretability enables a more informed discussion of the risks posed by advanced AI systems and lays firmer ground to engineer systems less likely to cause harm \citep{doshi2017towards,bender2021dangers,weidinger2022taxonomy,ngo2023alignment,carlsmith2023scheming}. 

% \td{define feature, use semicolors, directions extracted from the input}
\citet{olah2020zoom} propose three speculative claims regarding the interpretation of artificial neural networks: that features---directions in activation space representing properites of the input---are the fundamental unit of analysis, that features are connected into circuits via network weights, and that features and circuits are universal across models. That is, analogous features and circuits form in a diverse array of models and that different training trajectories converge on similar solutions \citep{li2015convergent}. Taken seriously, these hypotheses suggest a strategy for discovering important features and circuits: look for that which is universal. This line of reasoning motivates our work, where we leverage different notions of universality to identify and study individual neurons that represent features or underlie circuits. 

Beyond discovery, the degree to which neural mechanisms are universal is a basic open question that informs what kinds of interpretability research are most likely to be tractable and important. 
If the universality hypothesis is largely true in practice, we would expect detailed mechanistic analyses \citep{cammarata2021curve,wang2022interpretability,olsson2022context,nanda2023progress,mcdougall2023copy} to generalize across models such that it might be possible to develop a periodic table of neural circuits which can be automatically referenced when interpreting new models \citep{olah2020zoom}. Conversely, it becomes less sensible to dedicate substantial manual labor to understand low-level details of circuits if they are completely different in every model, and instead more efficient to allocate effort to engineering scalable and automated methods that can aid in understanding and monitoring higher-level representations of particular interest \citep{burns2022discovering,conmy2023towards,bills2023language,zou2023representation,bricken2023monosemanticity}. However, even in the case that not all features or circuits are universal, those which are common across models are likely to be more fundamental \citep{bau2018identifying,olsson2022context}, and studying them should be prioritized accordingly.

% Why do we care about universality / universal neurons
% \begin{itemize}
%     \item Tells us something about the degree of convergence/universality of neural representations.
%     \item Identifying that which is universal is likely to be the more fundamental units worth analyzing in more detail
%     \item Universal neurons are largely monosemantic, giving us a rough estimate of the number of monosemantic neurons in a model, potentially useful for estimating expansion sizes of feature dictionaries.
%     \item If monosemantic, gives us interpretable internal nodes of the network, ie, the inputs, outputs, or intermediate states of a circuit, to bootstrap further interpretability investigations.
% \end{itemize}
% we study universality at a more mechanistic level \citep{chughtai2023toy} as opposed to representational level \citep{kornblith2019similarity} more common in prior literature
In this work, we study the universality of individual neurons across GPT2 language models \citep{radford2019language} trained from five different random initializations \citep{Mistral}. While it is well known that individual neurons are often polysemantic \citep{nguyen2016multifaceted,olah2020zoom,elhage2022toy,gurnee2023finding} i.e., represent multiple unrelated concepts, we hypothesized that universal neurons were more likely to be monosemantic, potentially giving an approximation on the number of independently meaningful neurons. We choose to study models of the same architecture trained on the same data to have the most favorable experimental conditions for measuring universality to establish a rough bound for the universality over larger changes. We begin by operationalizing neuron universality in terms of activation correlations, that is, whether there exist pairs of neurons across different models which consistently activate on the same inputs. We compute pairwise correlations of neuron activations over 100 million tokens for every neuron pair across the different seeds and find that only 1-5\% of neurons pass a target threshold of universality compared to random baselines (\S~\ref{sec:number_universal}). We then study these universal neurons in detail, analyzing various statistical properties of both weights and activations (\S~\ref{sec:properties}), and find that they usually have clear interpretations and taxonomize them into a small number of neuron families (\S~\ref{sec:families}).
\begin{figure}
    \centering
    \begin{subfigure}{0.33\linewidth} 
        \includegraphics[width=\linewidth]{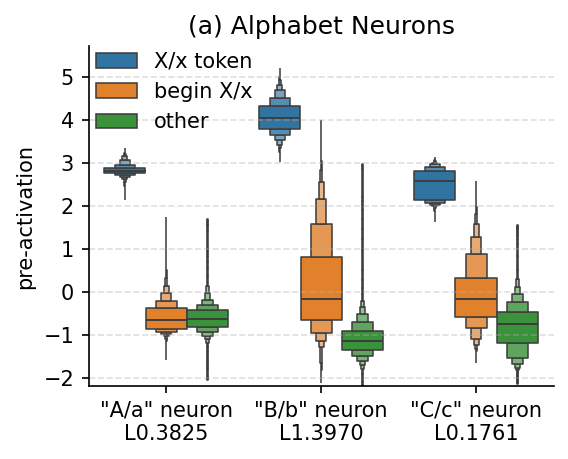} 
        \phantomcaption\label{fig:main_alphabet}
    \end{subfigure}
    \begin{subfigure}{0.33\linewidth} 
        \includegraphics[width=\linewidth]{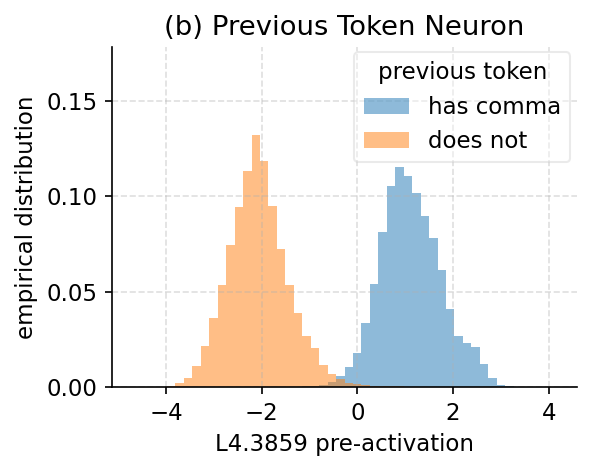} 
        \phantomcaption\label{fig:main_prev_token}
    \end{subfigure}
    \begin{subfigure}{0.33\linewidth} 
        \includegraphics[width=\linewidth]{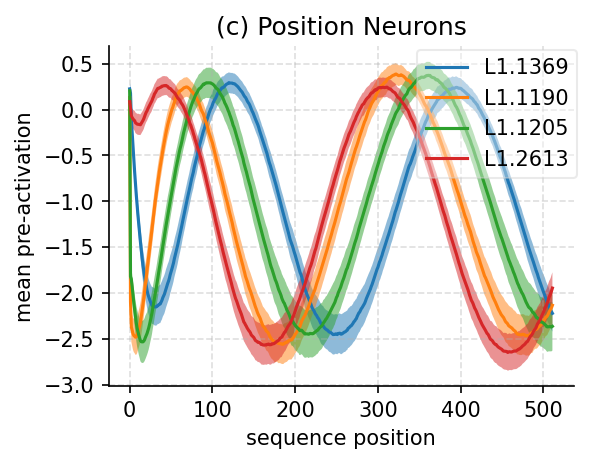} 
        \phantomcaption\label{fig:main_position}
    \end{subfigure}
    \begin{subfigure}{0.33\linewidth} 
        \includegraphics[width=\linewidth]{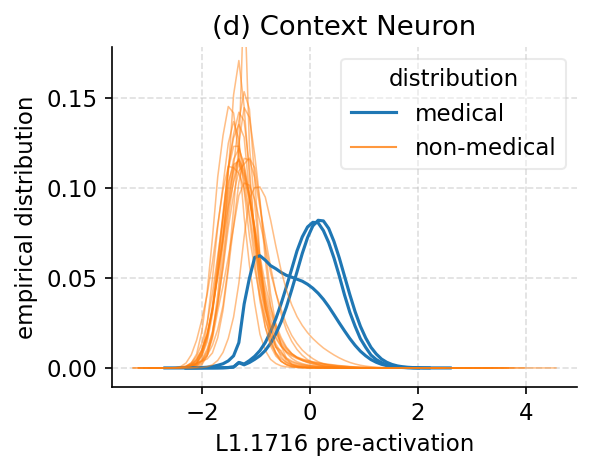} 
        \phantomcaption\label{fig:main_context}
    \end{subfigure}
    \begin{subfigure}{0.33\linewidth} 
        \includegraphics[width=\linewidth]{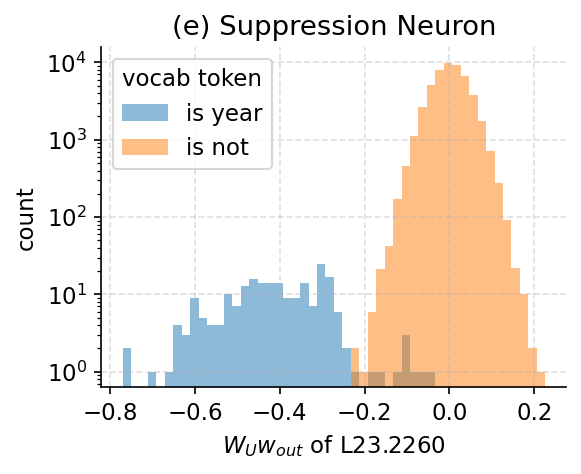} 
        \phantomcaption\label{fig:main_suppresion}
    \end{subfigure}
    \begin{subfigure}{0.33\linewidth} 
        \includegraphics[width=\linewidth]{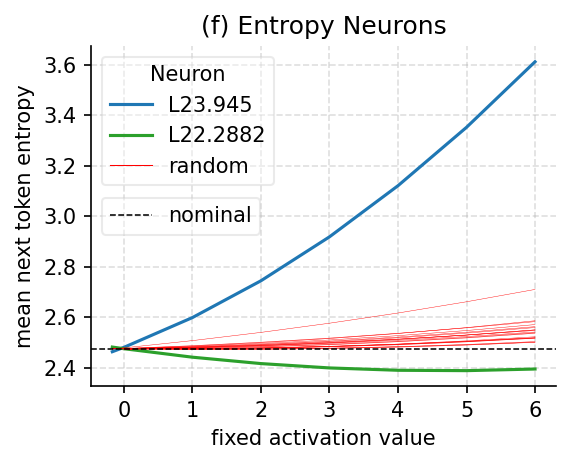} 
        \phantomcaption\label{fig:main_entropy_f}
    \end{subfigure}
    \caption{Universal neurons in GPT2 models, interpreted via their activations (a-d), weights (e), and causal interventions (f). 
    (a) Neurons which activate primarily on a specific individual letter and secondarily on tokens which begin with the letter; (b) Neuron which activates approximately if and only if the previous token contains a comma; (c) Neurons which activate as a function of absolute token position in the context (shaded area denotes standard deviation around the mean); (d) A neuron which activates in medical contexts (e.g. pubmed abstracts) but not in non-medical distributions; (e) a neuron which decreases the probability of predicting any integer tokens between 1700 and 2050 (i.e., years); (f) Neurons which change the entropy of the next token distribution when causally intervened.
    }
    \label{fig:main}
    \vspace{-3em}
\end{figure}

In Section \ref{sec:functional} we study a more abstract form of universality in terms of neuron weights rather than activations. That is, rather than understand a neuron in terms of the inputs which cause it to activate, understand a neuron in terms of the effects the neuron has on later model components or directly on the final prediction. Specifically, we analyze patterns in the compositional structure of the weights and find consistent outliers in how neurons affect other network components, constituting very simple circuits. In Section~\ref{sec:prediction}, we show there exists a large family of late layer neurons which have clear roles in predicting or suppressing a coherent set of tokens (e.g., second-person pronouns or single digit numbers), where the suppression neurons typically come in later layers than the prediction neurons. We then investigate a small set of neurons that leverage the final layer-norm operation to modulate the overall entropy of the next token prediction distribution (\S~\ref{sec:entropy}). We conclude with an analysis of neurons which control the extent to which an attention head attends to the first token, which empirically controls the output norm of the head, effectively turning a head on or off (\S~\ref{sec:attention}).

% What we do, mechanistic similarity instead of representational similarity
% - Run correlations
% - Look for patterns in the weights

% In summary, our main findings are that
% \begin{itemize}
%     \item Only 1-5\% of neurons are universal in the models we study, but these neurons are usually highly interpretable and belong to a small number of neuron families (e.g., unigram neurons, position neurons, syntax neurons, previous token neurons).
%     \item We characterize several universal functional roles of neurons
% \end{itemize}

\section{Related Work}

\paragraph{Universal Neural Mechanisms}
Features and circuits like high-low frequency detectors \citep{schubert2021high-low} and curve circuits \citep{cammarata2021curve} have been found to reoccur in vision models, with some features even reappearing in biological neural networks \citep{goh2021multimodal}. In language models, recent research has found similarly universal circuits and components like
induction heads \citep{olsson2022context} and successor heads \citep{gould2023successor}
and that models reuse certain circuit components to implement different tasks \citep{merullo2023circuit}. There has also been a flurry of recent work on studying more abstract universal mechanisms in language models like function vectors \citep{todd2023function,hendel2023context}, variable binding mechanisms \citep{feng2023language}, and long context retrieval \citep{variengien2023look}. Studying universality in toy models has provided ``mixed evidence'' on the universality hypothesis \citep{chughtai2023toy} and shown that multiple algorithms exist to implement the same tasks \citep{zhong2023clock,liao2023generating}.

% \td{Cite more}:
% successor heads \citep{gould2023successor}
% sentiment \citep{tigges2023linear}
% long context retrieval\citep{variengien2023look}
% Function vectors \citep{todd2023function,hendel2023context}
% Binding mechanisms \citep{feng2023language}

% Universality in brain \citep{goh2021multimodal}

\paragraph{Representational Similarity}
Preceding the statement of the universality hypothesis in mechanistic interpretability, there has been substantial work measuring representational similarity \citep{klabunde_similarity_2023}.
Common methods include canonical correlation analysis-based measures \citep{raghu_svcca_2017,morcos_insights_2018}, alignment-based measures \citep{hamilton_diachronic_2018,ding_grounding_2021,williams_generalized_2022,duong_representational_2023}, matrix-based measures \citep{kornblith2019similarity,tang_similarity_2020,shahbazi_using_2021,lin_geometric_2022,boix-adsera_gulp_2022,godfrey_symmetries_2023}, neighborhood-based measures \citep{hryniowski_inter-layer_2020,gwilliam_beyond_2022}, topology-based measures \citep{khrulkov_geometry_2018,barannikov_representation_2022}, and descriptive statistics \citep{wang_understanding_2022,lu_understanding_2022,lange_clustering_2022}. 
Previous work, mostly in vision models, has yielded mixed conclusions on whether networks with the same architecture learn similar representations. Some studies have found that networks with different initializations ``exhibit very low similarity'' \citep{wang_towards_2018} and ``do not converge to a unique basis'' \citep{brown2023privileged}, while others have shown that networks learn the same low-dimensional subspaces but not identical basis vectors \citep{li_convergent_2016} and that different models can be linearly stitched together with minimal loss suggesting they learn similar representations \citep{bansal_revisiting_2021}.

\paragraph{Analyzing Individual Neurons}
Many prior interpretability studies have analyzed individual neurons. In vision models, researchers have found neurons which activate for specific objects \citep{bau2020units}, curves at specific orientations \citep{cammarata2021curve}, high frequency boundaries \citep{schubert2021high},
multimodal concepts \citep{goh2021multimodal}, as well as for facets \citep{nguyen2016multifaceted} and compositions \citep{mu2020compositional} thereof. Moreover, many of these neurons seem universal across models \cite{dravid2023rosetta}. In language models, neurons have been found to correspond to sentiment \citep{radford2017learning,donnelly2019interpretability}, knowledge \citep{dai2021knowledge}, skills \citep{wang2022finding}, de-/re-tokenization \citep{elhage2022solu}, contexts \citep{gurnee2023finding,bills2023language}, position \citep{voita2023neurons}, space and time \citep{gurnee2023language}, and many other linguistic and grammatical features \citep{bau2018identifying,xin2019part,dalvi2019one,dalvi2020analyzing,durrani2022linguistic,sajjad2022analyzing}. More generally, it is hypothesized that neurons in language models form key-value stores \citep{geva2020transformer} that facilitate next token prediction by promoting concepts in the vocabulary space \citep{geva2022transformer}. However, many challenges exist in studying individual neurons, especially in drawing causal conclusions \citep{antverg2021pitfalls,huang2023rigorously}.

\section{Conceptual and Empirical Preliminaries} \label{sec:prelim}

\subsection{Universality}

% Model
% - Random seed
% - Model size
% - Hyperparameters
% - Architecture

% Data
% - Pretraining data (order and contents)

% Training
% - Loss function (regularization)
% - Finetuning

\paragraph{Notions of Universality}
Universality can refer to many different notions of similarity, each at a different level of abstraction and with differing measures and methodologies. Similar to Marr's levels of analysis in neuroscience \citep{hamrick2020levels,marr2010vision}, relevant notions of universality are: \textit{computational} or \textit{functional} universality regarding whether a (sub)network implements a particular input-output-behavior (e.g., whether the next token predictions for two different networks are the same); \textit{algorithmic} universality regarding whether or not a particular function is implemented using the same computational steps (e.g., whether a transformer trained to sort strings always learns the same sorting algorithm); \textit{representational} universality, or the degree of similarity of the information contained within different representations \citep{kornblith2019similarity} (e.g., whether every network represents absolute position in the context); and finally \textit{implementation} universality, i.e., whether individual model components learned by different models implement the same specialized computations (e.g., induction heads \citep{olsson2022context}, successor heads \citep{gould2023successor}, French neurons \citep{gurnee2023finding}, \textit{inter alia}). None of these notions of universality are usually binary, and the universality between components or computations can range from being formally isomorphic to simply sharing a common high-level conceptual or statistical motif.

In this work, we are primarily concerned with implementation universality in the form of whether individual neurons learn to specialize and activate for the same inputs across models. If such universal neurons do exist, then this is also a simple form of functional universality, as the distinct neurons constitute the final node of distinct subnetworks which compute the same output.

% - feature universality - can recover some feature from a model and is same across models
% - circuit universality - same algorithm
% - implementation level (monosemantic versus distributed)
% - Note you also do functional similarity
% - maybe add examples

\paragraph{Dimensions of Variations} Universality must be measured over some independent dimension of variation, that is, some change in the model, data or, training. For example, model variables include random seed, model size, hyperparameters, and architectural changes; data variables include the data size, ordering, and distribution of contents; training variables include loss function, optimizer, regularization, finetuning, and hyperparameters thereof. Assuming that changing random seed is the smallest change, this work primarily focuses on initialization universality in an attempt to bound the degree of similarity expected when studying larger changes.

% - Mechanistic universality (model components)
% - Representational universality (same information recoverable)
% - Algorithmic Universality (same algorithm implemented)
% - Functional Universality (same function implemented)
% - Statistical

\subsection{Models}

We restrict our scope to transformer-based auto-regressive language models \citep{radford2018improving} that currently power the most capable AI systems \citep{bubeck2023sparks}. Given an input sequence of tokens $x = [x_1, \dots, x_t] \in \mathcal{X} \subseteq \mathcal{V}^t$ from the vocabulary $\mathcal{V}$, a language model $\mathcal{M}: \mathcal{X} \rightarrow \mathcal{Y}$ outputs a probability distribution over the vocabulary to predict the next token in the sequence.

We focus on a replication of the GPT2 series of models  \citep{radford2019language} with some supporting experiments on the Pythia family \citep{biderman2023pythia}. For a GPT2-small and GPT2-medium architecture (see \S~\ref{sec:hparams} for hyperparameters) we study five models trained from different random seeds, referred to as GPT2-\{small, medium\}-[a-e] \citep{Mistral}.

\paragraph{Anatomy of a Neuron}
Of particular importance to this investigation is the functional form of the neurons in the feed forward (also known as multi-layer perceptron (MLP)) layers in the transformer. The output of an MLP layer given a normalized hidden state $\mathbf{x} \in \mathbb{R}^{d_\text{model}}$ is
\begin{equation} \label{eq:mlp}
    \textnormal{MLP}(\mathbf{x}) = \mathbf{W}_\text{out} \sigma(\mathbf{W}_\text{in} \mathbf{x} + \mathbf{b}_\text{in}) + \mathbf{b}_\text{out}
\end{equation}
where $\mathbf{W}^T_\text{out}, \mathbf{W}_\text{in} \in \mathbb{R}^{d_\text{mlp} \times d_\text{model}}$ are learned weight matrices, $\mathbf{b}_\text{in}$ and $\mathbf{b}_\text{out}$ are learned biases, and $\sigma$ is an elementwise nonlinear activation function. For all models we study, $\sigma$ is the GeLU activation function $\sigma(\mathbf{x}) = \mathbf{x}\Phi(\mathbf{x})$ \citep{hendrycks2016gaussian}. One can analyze an individual neuron $j$ in terms of the its activation $\sigma(\mathbf{w}^j_\text{in} \mathbf{x} + b^j_\text{in})$ for different inputs $\mathbf{x}$, or its weights---row $j$ of $\mathbf{W}_\text{in}$ or $\mathbf{W}_\text{out}^T$ which respectively dictate for what inputs a neuron activates and what effects it has downstream. 

We refer the reader to \citep{elhage2021mathematical} for a full description of the transformer architecture. We employ standard weight preprocessing techniques described further in \ref{sec:weight_preprocessing}.

\section{The Search for Universal Neurons} \label{sec:activation_universality}
\subsection{How Universal are Individual Neurons?} \label{sec:number_universal}
\paragraph{Experiment}
Inspired by prior work studying common neurons in neural networks \citep{li2015convergent,bau2018identifying,dravid2023rosetta}, we compute maximum pairwise correlations of neuron activations across five different models GPT2-\{a, b, c, d, e\} to find pairs of neurons across models which activate on the same inputs. In particular, let $N(a)$ be the set of neurons in model $a$. For each neuron $i \in N(a)$, 
we compute the Pearson correlation 
\begin{equation} \label{eq:pearson}
    \rho^{a, m}_{i, j} = \frac{\mathbb{E}\left[(\mathbf{v}^i - \mu_i)(\mathbf{v}^j - \mu_j) \right]}{\sigma_i \sigma_j}
\end{equation}

with all neurons $j \in N(m)$ in every model $m \in$\{b, c, d, e\}, where $\mu_i$ and $\sigma_i$ are the mean and standard deviation of a vector of neuron activations $\mathbf{v}^i$ computed across a dataset of 100 million tokens from the Pile test set \citep{gao2020pile}. For a baseline, we also compute $\bar{\rho}^{a, m}_{i, j}$, where instead of taking the correlation of $\rho(\mathbf{v}^i, \mathbf{v}^j)$, we compute $\rho(\mathbf{v}^i, (\mathbf{R}\mathbf{V})_j)$ for a random $d_\text{mlp} \times d_\text{mlp}$ Gaussian matrix $\mathbf{R}$ and the matrix of activations $\mathbf{V}$ for all neurons in a particular layer $N_\ell(m)$. In other words, we compute the correlation between neurons and elements within a random (approximate) rotation of a layer of neurons to establish a baseline correlation for the case where there does not exist a privileged basis \citep{elhage2021mathematical,brown2023privileged} to verify the importance of the neuron basis.

For a set of models $M$ we define the \textit{excess correlation} of neuron $i$ as the difference between the mean maximum correlation across models and the mean maximum baseline correlation in the rotated basis:
\begin{equation} \label{eq:excess}
   \varrho_i = \frac{1}{|M|}\sum_{m \in M} \left(\max_{j \in N(m)} \rho^{a, m}_{i, j} -  \max_{j \in N_R(m)} \bar{\rho}^{a, m}_{i, j}\right)
\end{equation}

\begin{figure}
    \centering
    \includegraphics[width=\linewidth]{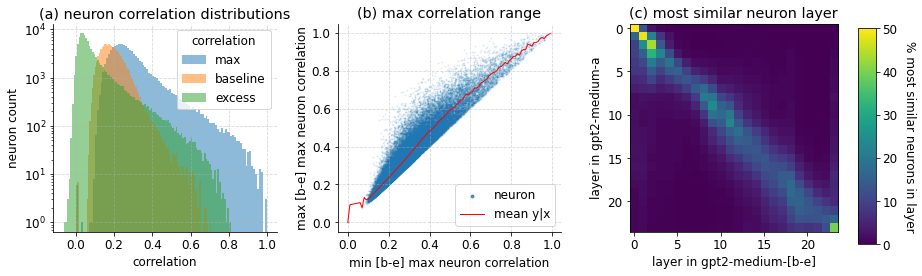}
    \caption{Summary of neuron correlation experiments in GPT2-medium-a. (a) Distribution of the mean (over models b-e) max (over neurons) correlation, the mean baseline correlation, and the difference (excess). (b) The max (over models) max (over neurons) correlation compared to the min (over models) max (over neuron) correlation for each neuron. (c) Percentage of layer pairs with most similar neuron pairs.}
    \label{fig:corr_summary_main}
\end{figure}

\paragraph{Results} Figure~\ref{fig:corr_summary_main} summarizes our results. In Figure~\ref{fig:corr_summary_main}a, we depict the average of the maximum neuron correlations across models [b-e], the average of the baseline correlations, and the excess correlation i.e., the left term, the right term, and the difference in (\ref{eq:excess}). While there is no principled threshold at which a neuron should be deemed universal, only 1253 out of the 98304 neurons in GPT2-medium-a have an excess correlation greater than 0.5.

To understand if high (low) correlation in one model implies high (low) correlation in all the models, in Figure~\ref{fig:corr_summary_main}b we report $\max_{m} \max_{j \in N(x)} \rho^{a, m}_{i, j}$ compared to $\min_{m} \max_{j \in N(m)} \rho^{a, m}_{i, j}$ for every neuron $i \in N(a)$. Figure~\ref{fig:corr_summary_main}b suggests there is relatively little variation in the correlations, as the mean difference between the max-max and min-max correlation is 0.049 for all neurons and 0.105 for neurons with $\varrho > 0.5$. Another natural hypothesis is that neurons specialize into roles based on how deep they are within the network (as suggested by \citep{olah2020zoom,elhage2022solu}). In \ref{fig:corr_summary_main}c, for each layer $l$ of model $a$, we compute the fraction of neurons in layer $l$ that have their most correlated neuron in layer $l^\prime$ for all $\ell^\prime$ in models [b-e]. Averaging across the different models, we observe significant \textit{depth specialization}, suggesting that neurons do perform depth specific computations, which we explore further in \S~\ref{sec:families}.

We repeat these experiments on GPT2-small and Pythia-160m displayed in Figures \ref{fig:corr_summary_small} and \ref{fig:corr_summary_pythia} respectively. A rather surprising finding is that while the percentage of universal neurons ($\varrho_i > 0.5$) within GPT2-medium and Pythia-160M are quite consistent (1.23\% and 1.26\% respectively), the number in GPT2-small-a is far higher at 4.16\%. We offer additional results and speculations in \S~\ref{sec:mysteries_scale}.

\subsection{Properties of Universal Neurons} \label{sec:properties}

We now seek to understand whether there are statistical proprieties associated with whether a neuron is universal or not, defined as having an excess correlation $\varrho_i > 0.5$. For all neurons in GPT2-medium-a, GPT2-small-a, and Pythia-160m, we compute various summary statistics of their weights and activations. For activations, we compute the mean, skew, and kurtosis of the pre-activation distribution over 100 million tokens, as well as the fraction of activations greater than zero, termed activation sparsity. For weights, we record the input bias $\mathbf{b}_\text{in}$, the cosine similarity between the input and output weight $\cos(\mathbf{w}_\text{in}, \mathbf{w}_\text{out})$, the weight decay penalty $\|\mathbf{w}_\text{in}\|_2^2 + \|\mathbf{w}_\text{out}\|_2^2$, and the kurtosis of the neuron output weights with the unembedding $\text{kurt}(\cos(\mathbf{w}_\text{out}, \mathbf{w}_{U}))$.

\begin{figure}
    \centering
    \includegraphics[width=\linewidth]{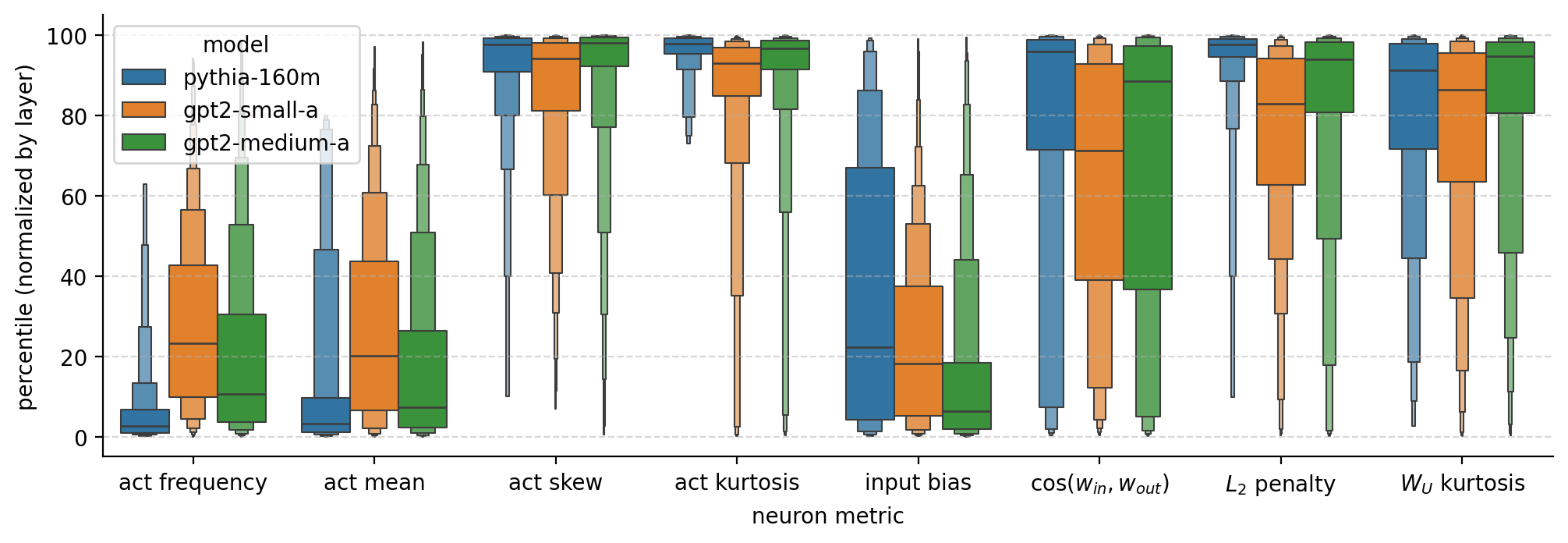}
    \caption{Properties of activations and weights of universal neurons for three different models, plotted as a percentile compared to neurons in the same layer.}
    \label{fig:universal_properties}
\end{figure}

In Figure~\ref{fig:universal_properties}, we report these statistics for universal neurons as a percentile compared to all neurons within the same layer; we choose this normalization to enable comparison across different layers, models, and metrics (a breakdown per metric and layer for GPT2-medium-a is given in Figure~\ref{fig:properties_by_layer}). Our results show that universal neurons do stand out compared to non-universal neurons. Specifically, universal neurons typically have large weight norm (implying they are important because the model was trained with weight decay) and have a large negative input bias, resulting in a large negative pre-activation mean and hence lower activation frequency. Furthermore, universal neurons have very high pre-activation skew and kurtosis, implying they usually have negative activation, but occasionally have very positive activation, proprieties we would expect of monosemantic neurons \citep{olah2020zoom,elhage2022toy,gurnee2023finding} which only activate when a specific feature is present in the input. In contrast, non-universal neurons usually have skew approximately 0 and kurtosis approximately 3, identical to a Gaussian distribution. We will discuss the meaning of high $\mathbf{W}_U$ kurtosis in \S~\ref{sec:prediction} and high $\cos(\mathbf{w}_\text{in}, \mathbf{w}_\text{out})$ in \S~\ref{sec:mysteries}.

\subsection{Universal Neuron Families} \label{sec:families}
Motivated by the observation that universal neurons have distributional statistics suggestive of monosemanticity, we zoom-in on individual neurons with $\varrho > 0.5$ and attempt to group them into a partial taxonimization of neuron families \citep{olah2020an,cammarata2021curve}. After manually inspecting many such neurons, we developed several hundred automated tests to classify neurons using algorithmically generated labels derived from elements of the vocabulary (e.g., whether a token \texttt{is\_all\_caps} or \texttt{contains\_digit}) and from the NLP package spaCy \citep{honnibal2020spacy}. Specifically, for each neuron with activation vector $\mathbf{v}$, and each test explanation which is a binary vector $\mathbf{y}$ over all tokens in the input, we compute the reduction in variance when conditioned on the explanation:
\begin{equation} \label{eq:reduction_in_variance}
1 - \frac{(1-\beta) \sigma^2(\mathbf{v} | \mathbf{y} = 0)  + \beta \sigma^2(\mathbf{v} | \mathbf{y} = 1)}{ \sigma^2(\mathbf{v})}
\end{equation}
where $\beta$ is the fraction of positive labels and $\sigma^2(\cdot)$ is the variance of a vector or subset thereof. Below, we qualitatively describe the most common families, and find our results replicate many findings previously documented in the literature.

\begin{figure}
    \centering
    \subcaptionbox{Near duplicate ``on'' unigram neurons \label{fig:unigram_main}}{\includegraphics[width=0.33\textwidth]{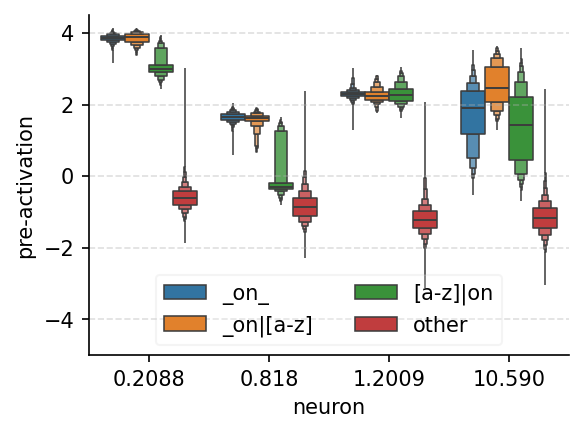}}
    \subcaptionbox{Syntax neuron \label{fig:syntax_main}}{\includegraphics[width=0.33\textwidth]{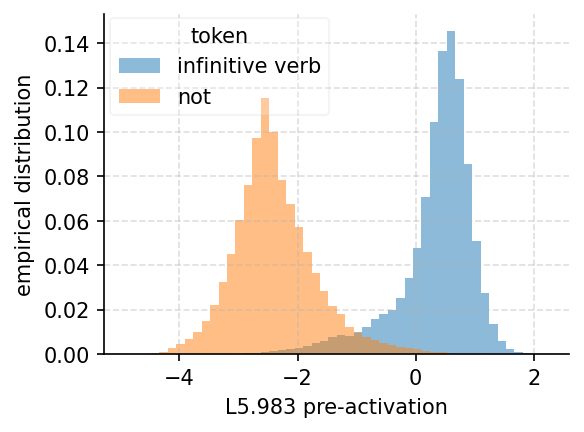}}
    \subcaptionbox{Place Neurons \label{fig:semantic_main}}{\includegraphics[width=0.33\textwidth]{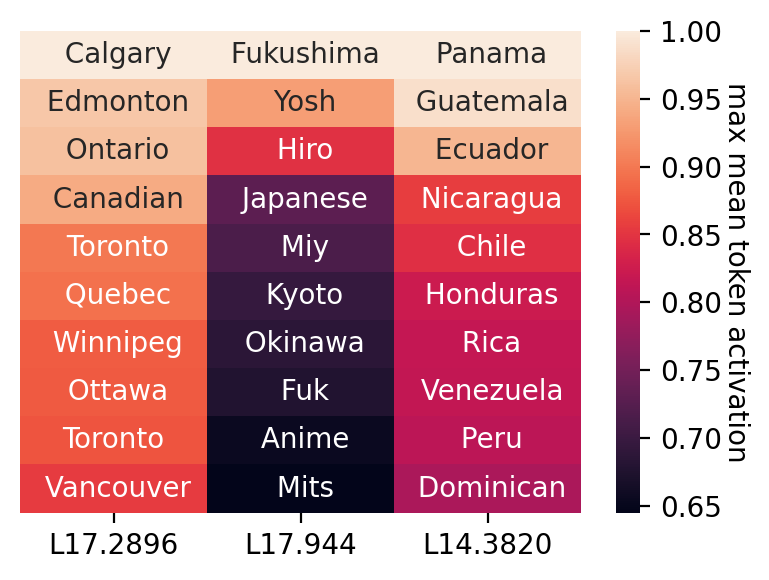}}
    \caption{Additional examples of universal neuron families in GPT2-medium.}
    \label{fig:additional_families}
\end{figure}

\paragraph{Unigram Neurons}
% \td{fix legend (remove +); naively this is surprising because could ahve all information in embedding, practioners need to be careful because embedding does not contain all information}
The most common type of neuron we found were \textit{unigram} neurons, which simply activate approximately if and only if the current token is a particular word or part of a word. These neurons often have many near duplicate neurons activating for the same unigram (Figure~\ref{fig:unigram_duplicates}) and appear predominately in the first two layers (Figure~\ref{fig:fig:unigram_nonduplicate}). One subtlety is the fact that common words will often have four different tokenizations from different combinations of capitalization and preceding space (e.g., ``\_on'' ``\_On'' ``on'' and ``On''). Therefore, for neurons responding to alphabetical unigrams, we breakdown activations depending on whether the unigram appears as a word, at the beginning of a word, or in the middle of a word (Figure~\ref{fig:unigram_main}), and find both positive and negative cases where the duplicate neurons respond to the unigram variations differently (Figures~\ref{fig:unigram_main} and \ref{fig:unigram_duplicates}). Such neurons illustrate that the token (un)embeddings may not contain all of the relevant token-level information, and that the model uses neurons to create an ``extended'' embedding of higher capacity.

\paragraph{Alphabet Neurons}
A particularly fun subclass of unigram neurons are \textit{alphabet} neurons (Figure~\ref{fig:main_alphabet}), which activate most strongly on tokens corresponding to an individual letter, and secondarily on tokens which begin with the respective letter. For 18 of 26 English letters there exist alphabet neurons with $\varrho > 0.5$ (Figure~\ref{fig:alphabet_neurons_full}), with some letters also having several near duplicate neurons.

\paragraph{Previous Token Neurons}
After finding an example of one neuron which seemed to activate purely as a function of the \textit{previous} token (e.g., if it contains a comma; Figure~\ref{fig:main_prev_token}), we decided to rerun our unigram tests with the labels shifted by one---that is, with the label given by the previous token. These tests surfaced many more previous token neurons occurring most often in layers 4-6 (see Figure~\ref{fig:prev_token_neurons_full} for an additional 25 universal previous token neurons). Such neurons illustrate the many potentially redundant paths of computations that can occur which complicates ablation based interpretability studies.

\paragraph{Position Neurons}
Inspired by the recent work of \citep{voita2023neurons}, we also run evaluations for \textit{position neurons}, neurons which activate as a function of absolute position rather than token or context (Figure~\ref{fig:main_position}). We follow the procedure of \citep{voita2023neurons} (who run their experiments on OPT models with ReLU activation \citep{zhang2022opt}) by computing the mutual information between activation and context position, and find similar results, with neurons that have a variety of positional patterns concentrated in layers 0-2 (see Figure~\ref{fig:position_appendix} for 20 more neurons). Similar to the unigram neurons, the presence of these neurons is potentially unexpected given their outputs could be learned directly by the positional embedding at the beginning of the model with less variance in activation.

\paragraph{Syntax Neurons}
Using the NLP package spaCy \citep{honnibal2020spacy}, we label our input data with part-of-speech, dependency role, and morphological data. We find many individual neurons that selectively activate for basic linguistic features like negation, plurals, and verb forms (Figure~\ref{fig:syntax_main}) which are not concentrated to any part of the network and resemble past findings on linguistic properties \citep{dalvi2019one,durrani2022linguistic}. Figure~\ref{fig:syntax_neurons_full} includes 25 more examples.

\paragraph{Semantic Neurons} Finally, we found a large number of neurons which activate for semantic features corresponding to coherent topics \citep{lim2023disentangling}, concepts \citep{elhage2022solu}, or contexts \citep{gurnee2023finding}. Such features are naturally much harder to algorithmically supervise. We use the subdistribution label from the Pile dataset \citep{gao2020pile} and manually labeled topics from an SVD based topic model as a best attempt, but this leaves many interpretable neurons undiscovered and uncategorized. In Figure~\ref{fig:semantic_main}, we show three regions neurons which activate most strongly on tokens corresponding to places in Canada, Japan, or Latin America respectively. Figure~\ref{fig:context_appendix} depicts 30 additional context neurons which activate on specific subdistributions, with many neurons which always activate for non-english text.

\section{Universal Functional Roles of Neurons} \label{sec:functional}
While the previous discussion was primarily focused on analyzing the \textit{activations} of neurons, and by extension the features they represent, this section is dedicated to studying the \textit{weights} of neurons to better understand their downstream effects. The neurons in this section are examples of \textit{action} mechanisms \citep{anthropic2023update}---model components that are better thought of as implementing an action rather than purely extracting or representing a feature, analogous to motor neurons in neuroscience.

\subsection{Prediction Neurons} \label{sec:prediction}

A simple but effective method to understand weights is through logit attribution techniques \citep{nostalgebraist2020interpreting,geva2022transformer,dar2022analyzing}. In this case, we can approximate a neuron's effect on the final prediction logits by simply computing  the product between the unembedding matrix and a neuron output weight $\mathbf{W}_U \mathbf{w}_\text{out}$ and hence interpret the neuron based on how it promotes concepts in the vocabulary space \citep{geva2022transformer}.

\begin{figure}
    \centering
    \includegraphics[width=\linewidth]{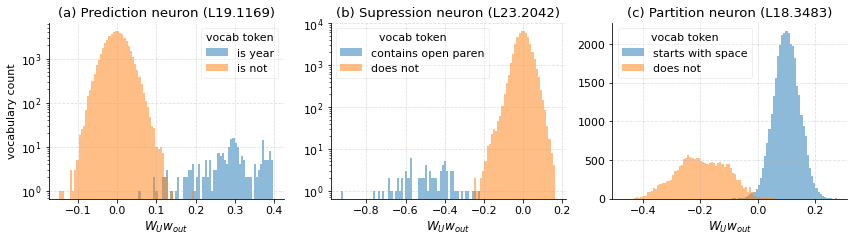}
    \caption{Example prediction neurons in GPT2-medium-a. Depicts the distribution of logit effects on the output vocabulary ($\mathbf{W}_U \mathbf{w}_\text{out}$) split by token property for 3 different neurons. (a) Prediction neuron increasing logits of integer tokens between 1700 and 2050 (i.e. years; high kurtosis), (b) Suppression neuron decreasing logits for tokens containing an open parenthesis (high kurtosis and negative skew), and (c) Partition neuron boosting tokens beginning with a space and suppressing tokens which do not (high variance; note, linear y-scale).}
    \label{fig:prediction_logits_main}
\end{figure}

When we apply our automated tests from \S~\ref{sec:families} on $\mathbf{W}_U \mathbf{w}_\text{out}$ rather than the activations for our universal neurons, we found several general patterns (Figure~\ref{fig:prediction_logits_main}), many individual neurons with extremely clear interpretations (Figure~\ref{fig:prediction_neurons_full}), and clusters of neurons which all affect the same tokens  (Figure~\ref{fig:prediction_neurons_duplicates}). Specifically, we find many examples of \textit{prediction} neurons that positively increase the predicted probability of a coherent set of tokens while leaving most others approximately unchanged (Fig~\ref{fig:prediction_logits_main}a); \textit{suppression} neurons that are similar, except decrease the probability of a group of related tokens (Fig~\ref{fig:prediction_logits_main}b); and \textit{partition} neurons that partition the vocabulary into two groups, increasing the probability of one while decreasing the probability of the other (Fig~\ref{fig:prediction_logits_main}c). The prediction, suppression, and partition motifs can be automatically detected by studying the moments of the distribution of vocabulary effects given by $\mathbf{W}_U\mathbf{w}_\text{out}$. In particular, both prediction and suppression neurons will have high kurtosis (the fourth moment---a measure of how much mass is in the tails of a distribution), but prediction neurons will have positive skew and suppression neurons will have negative skew. Partition neurons will shift the probability of most tokens and have high variance in overall logit effect. From this, we see almost all universal neurons ($\varrho > 0.5$) in later layers are one of these prediction neuron variants (Figure~\ref{fig:properties_by_layer}).

\begin{figure}
    \centering
    \includegraphics[width=\linewidth]{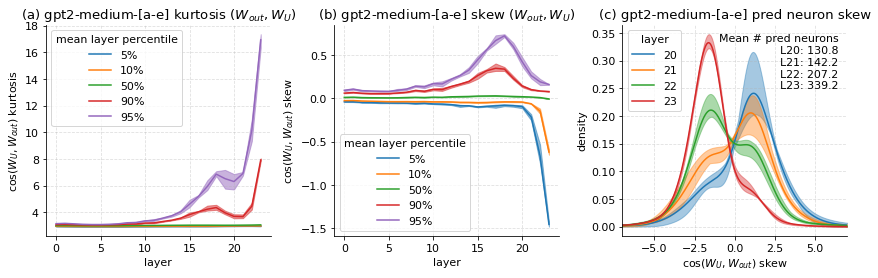}
    \caption{Summary statistics of cosine similarity between neuron output weights ($\mathbf{W}_\text{out}$) and token unembedding ($\mathbf{W}_{U}$) for GPT2-medium-[a-e]. (a,b) Percentiles of kurtosis and skew by layer averaged over [a-e]. (c) Distribution of skews for neurons with kurtosis greater than 10 in last four layers. Shaded area denotes range across all five models.}
    \label{fig:prediction_percentiles}
\end{figure}

To better understand the number and location of these prediction neurons, we compute the moment metrics of $\cos(\mathbf{W}_U, \mathbf{w}_\text{out})$ for all neurons in all five GPT2-medium models, and show how these statistics vary over model depth in Figure~\ref{fig:prediction_percentiles}. We find a striking pattern which is remarkably consistent across the different seeds: after about the halfway point in the model, prediction neurons become increasingly prevalent until the very end of the network where there is a sudden shift towards a much larger number of suppression neurons. To ensure this is not just an artifact of the tied embeddings ($\mathbf{W}_E$ = $\mathbf{W}_U^T$) in the GPT2 models, we also run this analysis on five Pythia models ranging from 410M to 6.9B parameters and find the results are largely the same (Figure~\ref{fig:pythia_logit_stats}).

When studying the activations of suppression neurons, we noticed that they activate far more often when the next token is in fact from the set of tokens they suppress (e.g., a year token like ``1970''; Figure~\ref{fig:prediction_neurons_duplicates}). We intuit that these suppression neurons fire when it is plausible but not certain that the next token is from the relevant set. Combined with the observation that there exist many suppression and prediction neurons for the same token class (Figure~\ref{fig:prediction_neurons_duplicates}), we take this as evidence of an ensemble hypothesis where the model uses multiple neurons with some independent error that combine to form a more robust and calibrated estimate of whether the next token is in fact a year.

In addition to being a clean example of an action mechanism \citep{anthropic2023update}, these results are interesting as they refine a conjecture made by \citep{geva2022transformer}. Specifically, rather than ``feed-forward layers build predictions by promoting concepts in the vocabulary space,'' we claim \textit{late} feed-forward (MLP) layers build predictions by both promoting \textit{and} suppressing concepts in the vocabulary space. Moreover, it suggests there are different stages in the iterative inference pipeline \citep{belrose2023eliciting,jastrzkebski2017residual}, where first affirmative predictions are made, and then the distribution is sharpened or made more calibrated by suppression neurons at the very end. The existence of suppression neurons also sheds light on recent observations of individual neurons \citep{bills2023language} and MLP layers \citep{mcgrath2023hydra} suppressing the maximum likelihood token and being a mechanism for self-repair.
% \td{That is, if you ablated an ealier component which caused suppresion, it would not supress, and undo the ablation}

\subsection{Entropy Neurons} \label{sec:entropy}

\begin{figure}
    \centering
    \includegraphics[width=\linewidth]{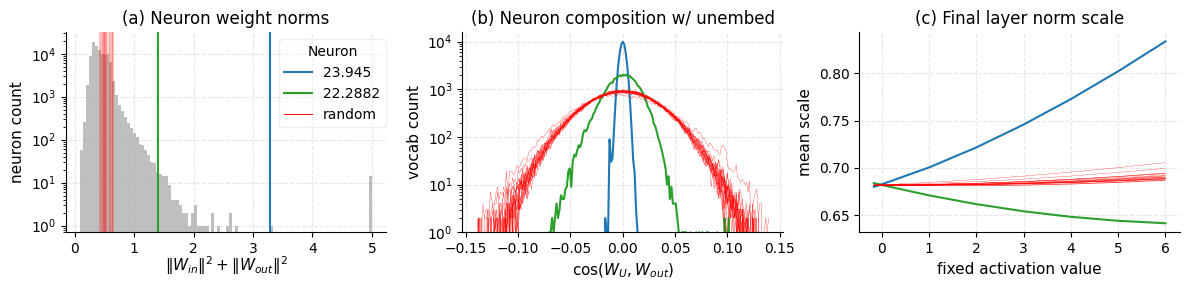}
    \includegraphics[width=\linewidth]{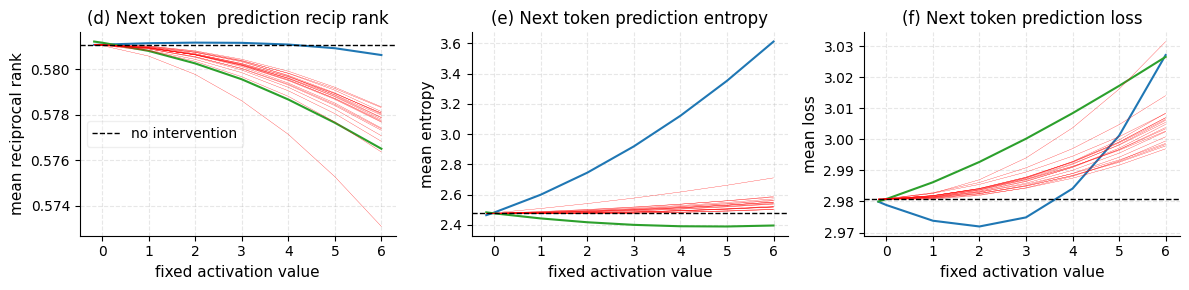}
    \caption{Summary of (anti-)entropy neurons in GPT2-medium-a compared to 20 random neurons from final two layers. Entropy neurons have high weight norm (a) with output weights mostly orthogonal to the unembedding matrix (b). Fixing the activation to larger values causes the final layer norm scale to increase dramatically (c) while leaving the ranking of the true next token prediction mostly unchanged (d). Increased layer norm scale squeezes the logit distribution, causing a large increase in the prediction entropy (e; or decrease for anti-entropy neuron) and an increase or decrease in the loss depending on the model's baseline level of under- or over-confidence (f). Legend applies to all subplots.}
    \label{fig:entropy_interventions_main}
\end{figure}

Because models are trained with weight decay ($\ell_2$ regularization) we hypothesized that neurons with large weight norms would be more interesting or important because they come at a higher cost. While most turned out to be relatively uninteresting (mostly neurons which activate for the beginning of sequence token), the $15^{\text{th}}$ largest norm neuron in GP2-medium-a (L23.945) had an especially interesting property: it had the lowest variance logit effect $\mathbf{W}_U \mathbf{w}_\text{out}$ of any neuron in the model; i.e., it only has a tiny effect on the logits. To understand why a final layer neuron, which can only affect the final logit distribution, has high weight norm while performing an approximate no-op on the logits, recall the final decoding formula for the probability of the next token given a final residual stream vector $\mathbf{x}$
\begin{equation} \label{eq: decoding}
    p(\mathbf{y} | \mathbf{x}) = \text{Softmax}(\mathbf{W}_U \text{LayerNorm}(\mathbf{x})), \quad \quad \text{LayerNorm}(\mathbf{x}) = \frac{\mathbf{x} - \mathbb{E}[\mathbf{x}]}{\sqrt{\text{Var}[\mathbf{x}] + \epsilon}}.
\end{equation}

% - Looking for neurons with large magntitude
% - Mostly BOS but the 15th largest had an interesting property: extremely low variance with the vocabulary
% - To understand what is going on recall the final decoding formula

We hypothesize that the function of this neuron is to modulate the model's uncertainty over the next token by using the layer norm to squeeze the logit distribution, in a manner quite similar to manually increasing the temperature when performing inference. To support this hypothesis, we perform a causal intervention, fixing the neuron in question to a particular value and studying the effect compared to 20 random neurons from the last two layers that are not in the top decile of norm or in the bottom decile of logit variance (Figure~\ref{fig:entropy_interventions_main}). We find that intervening on this \textit{entropy} neuron indeed causes the layer norm scale to increase dramatically (because of the large weight norm) while largely not affecting the relative ordering of the vocabulary (because of the low composition), having the effect of increasing overall entropy by dampening the post-layer norm component of $\mathbf{x}$ in the row space of $\mathbf{W}_U$.

Additionally, we observed a neuron (L22.2882) with $\cos(\mathbf{w}_\text{out}^{23.945}, \mathbf{w}_\text{out}^{22.2882}) = -0.886$ (i.e., a neuron that writes in the opposite direction forming an antipodal pair \citep{elhage2022toy}) that also has high weight norm. Repeating the intervention experiment, we find this neuron \textit{decreases} the layer norm scale and decreases the mean next token entropy, forming an anti-entropy neuron. These results suggest there may be one or more global uncertainty directions that the model maintains to modulate its overall confidence in its prediction. However, our experiments with fixed activation value do not necessarily imply the model uses these neurons to increase the entropy as a general uncertainty mechanism, and we did notice cases in which increasing the activation of the entropy neuron decreased entropy, suggesting the true mechanism may be more complicated.

We repeat these experiments on GPT2-small-a and find an even more dramatic antipodal pair of (anti-)entropy neurons  in Figure~\ref{fig:entropy_interventions_gpt2_small}. To our knowledge, this is the first documented mechanism for uncertainty quantification in language models and perhaps the second example of a mechanism involving layer norm \citep{brody2023expressivity}.

\subsection{Attention Deactivation Neurons} \label{sec:attention}

\begin{figure}
    \centering
    \includegraphics[width=\linewidth]{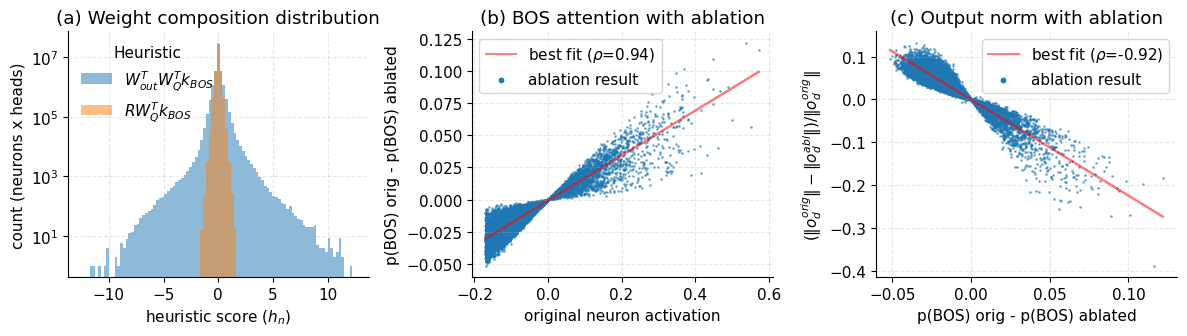}
    \caption{Summary of attention (de-)activation neuron results in GPT2-medium-a. (a) Distribution of heuristic score $h_n$ for every pair of neurons and heads compared to random neuron directions $\mathbf{R}$. (b;c) path ablations effect of neuron L4.3594 on head L5.H0: ablating positive activation reduces attention to BOS (b) causing the norm to increase (c). }
    \label{fig:deactivation}
\end{figure}

In autoregressive models, attention heads frequently place all of their attention on the beginning of sequence (BOS) token \citep{xiao2023efficient}. We hypothesise that the model uses the attention to the BOS token as a kind of (de-)activation for the head, where fully attending to BOS implies the head is deactivated and has minimal effect. Moreover, we hypothesize that there are individual neurons which control the extent to which heads attend to BOS. 

Recall the output of an attention head $\mathbf{o}_d$ for a destination token $d$ from source tokens $s$ is given by $$\mathbf{q}_d = \mathbf{W}_Q \mathbf{r}_d, \ \ \mathbf{k}_s = \mathbf{W}_K \mathbf{r}_s, \ \ \mathbf{S}_{ds} = \mathbf{q}_d^T \mathbf{k}_s, \ \  \mathbf{A}_{ds} = \softmax_s(\frac{M(\mathbf{S}_{ds})}{\sqrt{d_h}}), \ \  \mathbf{v}_s = \mathbf{W}_V \mathbf{r}_s , \ \  \mathbf{o}_d = \mathbf{W}_O \sum_s \mathbf{A}_{d s} \mathbf{v}_s $$ where $\mathbf{r}_{s / d}$ is the residual stream at the source / destination token, $d_h$ is the bottleneck dimension of the head, and $M(\cdot)$ applies the causal attention mask to the attention scores. The calculation of the attention pattern $\mathbf{A}_{ds}$ via a $\softmax$ across the source positions means that the attention given to the source tokens by a given destination token sums to one. 

% In causal models, attention patterns frequently place significant attention on the BOS token. We hypothesise that the model uses the logit from the BOS source token as a kind of activation for the head, allowing greater control over the norms of the outputs from other source tokens. 

The vector $\mathbf{W}_O \mathbf{v}_{BOS}$ is constant for all prompts and contains no semantic information. If it has a low norm, attending to BOS scales down the outputs of attending to other source positions while maintaining their relative attention because the attention scores must sum to one. If the BOS output norm is near zero, the head can effectively turn off by only attending to the BOS token. In practice, the median head in GPT-2-medium-a has a $\mathbf{W}_O \mathbf{v}_{BOS}$ with norm 19.4 times smaller than the average for other tokens.

We can identify neurons which may use this mechanism for a given head by a heuristic score $h_n = \mathbf{W}_{out}^T \mathbf{W}_Q^T \mathbf{k}_{BOS}$ for unit normalized $\mathbf{W}_{out}$. Positive scores suggests activation of the neuron will increase the attention placed on BOS, decreasing the output norm of the head, and the opposite for negative scores. Figure~\ref{fig:deactivation}a shows the distribution of the scores for all heads in GPT2-medium-a compared to a unit normalized Gaussian matrix $\mathbf{R}$. 
% We note that, for the middle to late layers in particular, we find that the tails of these distributions are significantly heavier than random.

% For a given neuron, we can measure the effect of activation on the output norm of a given head by path ablation \citep{wang2022interpretability} of the neuron at a given destination token and taking the difference in norm of the output of the head between the original and ablated runs. We perform this procedure on a large number of positions late in the context and also record the change in the attention placed on BOS token. Figure Figure~\ref{deactivation}b and \ref{deactivation}c take the highest scoring neuron the particular head shows plots relating the activation of the neuron, the corresponding BOS attention and head output norm differences. This is an example of an attention [] neuron.

For a given neuron, we can measure the effect of activation on the attention to BOS and output norm of a given head by path ablation \citep{wang2022interpretability} of the neuron at a particular destination token. Specifically, we can measure the difference in BOS attention and norm of the output of the head between the original run and a forward pass where the contribution of a neuron is deleted (i.e, zero path ablated) from the input of a particular head at the current token position. We perform this procedure over a random subset of tokens in the second half of the context to avoid spurious effects stemming from short contexts. Figure~\ref{fig:deactivation}b and \ref{fig:deactivation}c depict the results of these path ablations for the highest scoring neuron in layer 4 for head 0 in attention layer 5. This is an example of an attention deactivation neuron---increasing the activation of the neuron increases the attention to BOS reducing the output norm of the head $\|\mathbf{o}_d\|$. See Figure~\ref{fig:more_attention_activation_gpt2_medium} for 5 additional examples of attention (de-)activating neurons.

% We can see the degree to which these neurons pass through the BOS logit mechanism as opposed to others by performing our path ablations only on the attention attention scores with BOS as the source token and comparing the results to path ablation for the entire head. Figure () shows the distributions of the correlation coefficients and ratio of the standard deviation of the norm difference and neuron activation for MLP layer () and head () in attention layer () in GPT-2-medium-a.

\section{Additional Mysteries} \label{sec:mysteries}
We conclude our investigation by commenting on several miscellaneous results that we think are 
worth reporting but that we do not fully understand.

\subsection{Cosine and Activation Frequency} \label{sec:mysteries_cosine}
\begin{figure}
    \centering
    \includegraphics[width=\linewidth]{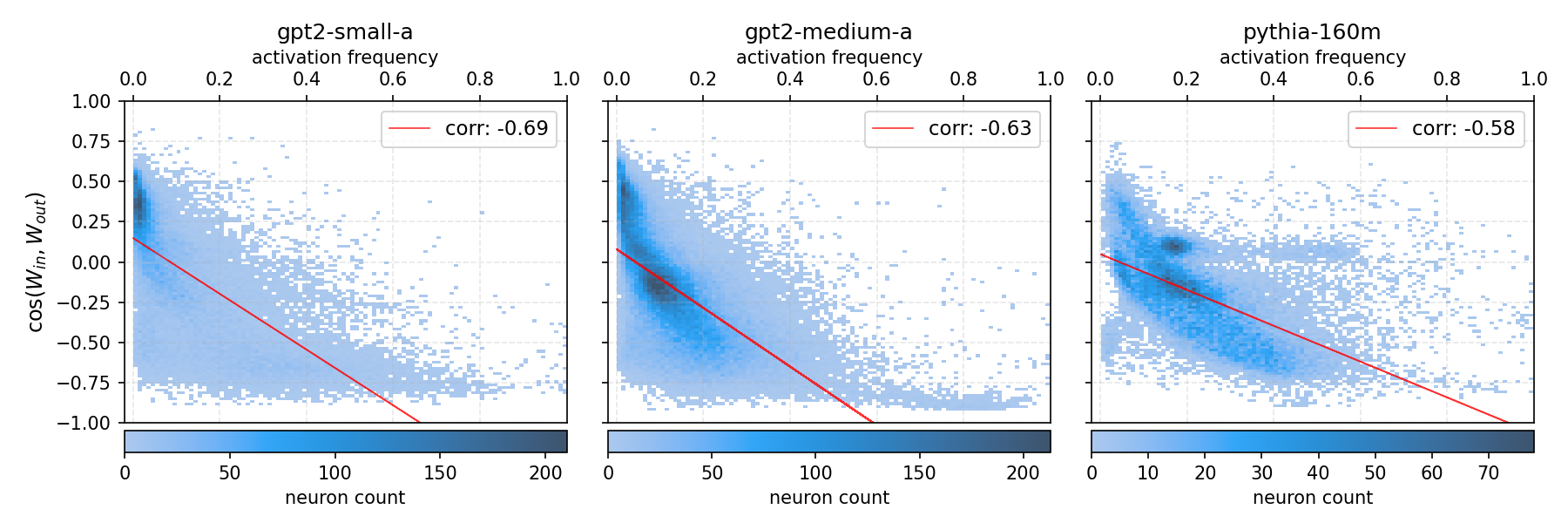}
    \caption{Activation frequency of neuron (fraction of activation values greater than zero) versus cosine similarity of neuron input and output weights for GPT2-small-a (left), GPT2-medium-a (center), and Pythia-160M (right).}
    \label{fig:sparsity_vs_cos}
\end{figure}

An unexpectedly strong relationship we observed is the correlation between activation frequency of a neuron and the cosine similarity between its input and output weight vectors $\cos(\mathbf{w}_\text{in}, \mathbf{w}_\text{out})$ as shown in Figure~\ref{fig:sparsity_vs_cos}. Almost all neurons with a very high activation frequency have input and output weights in almost opposite directions. These neurons are predominantly in the first quarter of network depth and have small excess correlation, i.e., they are not universal as measured by activation. We also find it noteworthy that there appears to be an approximate ceiling and floor on the cosine similarity of approximately $\pm$0.8.

\subsection{Duplication and Universality}\label{sec:mysteries_duplication}
\begin{figure}
    \centering
    \includegraphics[width=\linewidth]{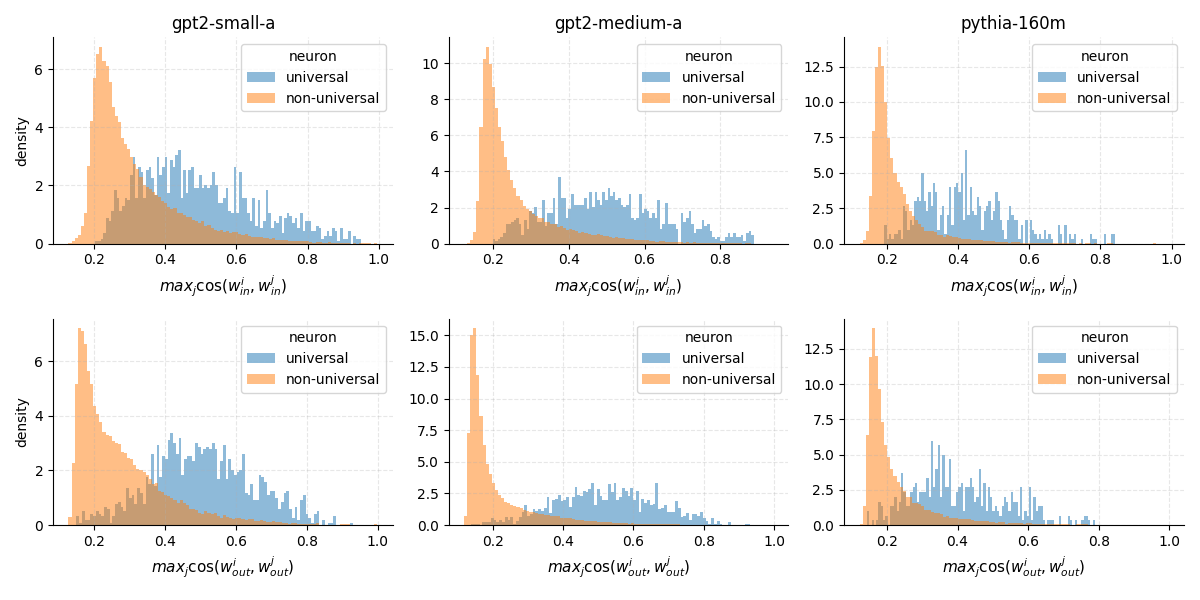}
    \caption{Distribution of cosine similarities of most similar neurons measured by input weights (top) and output weights (bottom) for GPT2-small-a (left), GPT2-medium-a (middle), and Pythia-160M (right) colored by universality ($\varrho > 0.5$).}
    \label{fig:max_weight_sim}
\end{figure}

% - Redundancy happens in weight space
% - reference figure
% -discuss hypotheses

% Se
% - Ensemble hypothesis
% - Instead of dead neurons
% - Vestigial
% - Dropout
% - Features in superposition with rare features

While neuron redundancy has been observed in models before \citep{casper2021frivolous,dalvi2020analyzing} and large models can be effectively pruned \citep{xia2023sheared}, we were surprised by the number of seemingly duplicate universal neurons we observed (e.g., Figure \ref{fig:unigram_duplicates} or the 105 BOS neurons we observed). Naively, this is surprising, as it seems wasteful to dedicate multiple neurons to the same feature. Larger models have more capacity and are empirically much more effective so why have redundant neurons when you could instead have one neuron with twice the output weight norm?

A few potential explanations are (1) these models were trained with weight decay, creating an incentive to spread out the computation. (2) Dropout---however, in these models dropout is applied to the output of the MLP layer, rather than the MLP activations themselves. (3) These neurons are vestigial remnants that were useful earlier in training \citep{quirke2023training}, but are potentially stuck in a local minima and are no longer useful. (4) The duplicated neurons are only activating the same on common features, but are polysemantic with different sets of rarer features. (5) Ensembling, where each neuron computes the same feature but with some independent error, and together form an ensemble with lower average error.

By measuring redundancy in terms of similarity in weights (Figure~\ref{fig:max_weight_sim}), we find very few neurons which are literal duplicates, providing more evidence for (4) and (5). Based on the much higher level of similarity for universal neurons, it is possible this effect is relatively small in general.

\subsection{Scale and Universality} \label{sec:mysteries_scale}

\begin{figure}
    \centering
    \includegraphics[width=\linewidth]{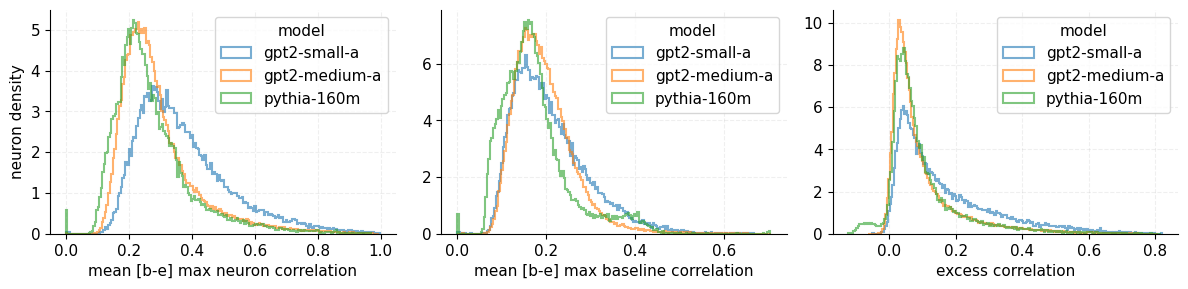}
    \caption{Empirical distribution of max neuron correlation averaged across models (left), max baseline correlation averaged across models (middle), and the difference denoted as the excess correlation (right). 
    }
    \label{fig:correlation density}
\end{figure}

As mentioned in \S~\ref{sec:activation_universality}, GPT2-medium and Pythia-160M have a consistent number of universal neurons (1.23\% and 1.26\% respectively), while GPT2-small-a has many more 4.16\%. In Figure~\ref{fig:correlation density} we show the distribution of max, baseline, and excess correlations for all models, where we see that GPT2-medium and Pythia-160M have almost identical distributions while GPT2-small is an outlier. GPT2-small also has correspondingly greater weight redundancy as shown in Figure~\ref{fig:max_weight_sim}. One explanation for this is the number of universal neurons decreases in larger models. This is potentially implied by results from \citep{bills2023language} who observe larger models have fewer neurons which admit high quality natural language interpretations. However, without additional experiments on larger models trained from random seeds, this remains an open question.

\section{Discussion and Conclusion}

\paragraph{Findings} In this work, we explore the universality of individual neurons in GPT2 language models, and find that only about 1-5\% of neurons are universal across models, constituting another piece of evidence that individual neurons are not the appropriate unit of analysis for most network behaviours. Nonetheless, we have shown that leveraging universality is an effective unsupervised approach to identify interpretable model components and important motifs. In particular, those few neurons which are universal are often interpretable, can be grouped into a smaller number of neuron families, and often develop with near duplicate neurons in the same model. Some universal neurons also have clear functional roles, like modulating the next token prediction entropy, controlling the output norm of an attention head, and predicting or suppressing elements of the vocabulary in the prediction. Moreover, these functional neurons often form antipodal pairs, potentially enabling collections of neurons to ensemble to improve robustness and calibration.

\paragraph{Limitations} Compared to frontier LLMs, we study small models of only hundreds of million parameters and tens of thousands of neurons due to the expense of training multiple large scale language models from different random initializations. We also study a relatively narrow form of universality: neuron universality over random seeds within the same model family. Studying universality across different model families is made difficult by tokenization discrepancies, and studying models across larger sizes is difficult due to the expense of computing all pairwise neuron correlations over a sufficiently sized text corpus. Additionally, many of our interpretations rely on manual analysis or algorithmic supervision which restricts the scope and generality of our methods. Moreover, our narrow focus on a subset of individual elements of the neuron basis potentially obscures important details and ignores the vast majority of overall network computation.

% \begin{itemize}
%     \item Small model size (due to expense of correlations and lack of large models trained from random seeds).
%     \item Can compare within model families but not across models families due to tokenization discrepancies.
%     \item Analysis is largely manual
%     \item Neuron basis is often times the wrong level of analysis
% \end{itemize}

\paragraph{Future Work} Each of these limitations suggest avenues for future work. Instead of studying the neuron basis, our experiments could be replicated on an overcomplete dictionary basis that is more likely to contain the true model features \citep{cunningham2023sparse,bricken2023monosemanticity}. Motivated by the finding that the most correlated neurons occur in similar network depths, our experiments could be rerun on larger models where pairwise correlations are only computed between adjacent layers to improve scalability. Additionally, the interpretation of common units could be further automated using LLMs to provide explanations \citep{bills2023language}. Finally, by uncovering interpretable footholds within the internals of the network, our findings can form the basis of deeper investigations into how these components respond to stimulus or perturbation, develop over training \citep{quirke2023training}, and affect downstream components to further elucidate general motifs and specific circuits within language models.

% \begin{itemize}
%     \item Apply techniques to feature dictionaries
%     \item Apply techniques to larger models (compute per layer to enable scaling)
%     \item Automate analysis
%     \item Use neurons as interpretable nodes of circuit
% \end{itemize}

\section*{Acknowledgments}
We would like to thank Yossi Gandelsman, Lovis Heindrich, Lucia Quirke, for useful discussions and comments on our work. We made extensive use of the TransformerLens library \citep{nandatransformerlens2022} and the MIT Supercloud \citep{reuther2018interactive} for our experiments and computational resources. WG was partially supported by an Open Philanthropy early career grant.

\section*{Author Contribution}
WG led the project, conducted most of the analysis, and wrote most of the paper. TH led the effort on understanding attention (de-)activation neurons, and performed the corresponding analysis and writing. ZCG assisted with experimental infrastructure. ZCG, TRK, QS, and WH assisted with neuron analysis and writing/editing. NN gave frequent and detailed feedback on experiment design and analysis in addition to editing the paper. DB supported the project and edited the paper.

%Bibliography
\bibliographystyle{apalike}  
\bibliography{references}  

\begin{thebibliography}{}

\bibitem[Anthropic, 2023]{anthropic2023update}
Anthropic (2023).
\newblock Circuits updates - july 2023.
\newblock https://transformer-circuits.pub/2023/july-update/index.html.

\bibitem[Antverg and Belinkov, 2021]{antverg2021pitfalls}
Antverg, O. and Belinkov, Y. (2021).
\newblock On the pitfalls of analyzing individual neurons in language models.
\newblock {\em arXiv preprint arXiv:2110.07483}.

\bibitem[Bansal et~al., 2021]{bansal_revisiting_2021}
Bansal, Y., Nakkiran, P., and Barak, B. (2021).
\newblock Revisiting {Model} {Stitching} to {Compare} {Neural}
  {Representations}.

\bibitem[Barannikov et~al., 2022]{barannikov_representation_2022}
Barannikov, S., Trofimov, I., Balabin, N., and Burnaev, E. (2022).
\newblock Representation {Topology} {Divergence}: {A} {Method} for {Comparing}
  {Neural} {Network} {Representations}.

\bibitem[Bau et~al., 2018]{bau2018identifying}
Bau, A., Belinkov, Y., Sajjad, H., Durrani, N., Dalvi, F., and Glass, J.
  (2018).
\newblock Identifying and controlling important neurons in neural machine
  translation.
\newblock {\em arXiv preprint arXiv:1811.01157}.

\bibitem[Bau et~al., 2020]{bau2020units}
Bau, D., Zhu, J.-Y., Strobelt, H., Lapedriza, A., Zhou, B., and Torralba, A.
  (2020).
\newblock Understanding the role of individual units in a deep neural network.
\newblock {\em Proceedings of the National Academy of Sciences}.

\bibitem[Belrose et~al., 2023]{belrose2023eliciting}
Belrose, N., Furman, Z., Smith, L., Halawi, D., Ostrovsky, I., McKinney, L.,
  Biderman, S., and Steinhardt, J. (2023).
\newblock Eliciting latent predictions from transformers with the tuned lens.
\newblock {\em arXiv preprint arXiv:2303.08112}.

\bibitem[Bender et~al., 2021]{bender2021dangers}
Bender, E.~M., Gebru, T., McMillan-Major, A., and Shmitchell, S. (2021).
\newblock On the dangers of stochastic parrots: Can language models be too big?
\newblock In {\em Proceedings of the 2021 ACM conference on fairness,
  accountability, and transparency}, pages 610--623.

\bibitem[Bengio et~al., 2023]{bengio2023managing}
Bengio, Y., Hinton, G., Yao, A., Song, D., Abbeel, P., Harari, Y.~N., Zhang,
  Y.-Q., Xue, L., Shalev-Shwartz, S., Hadfield, G., et~al. (2023).
\newblock Managing ai risks in an era of rapid progress.
\newblock {\em arXiv preprint arXiv:2310.17688}.

\bibitem[Biderman et~al., 2023]{biderman2023pythia}
Biderman, S., Schoelkopf, H., Anthony, Q., Bradley, H., O'Brien, K., Hallahan,
  E., Khan, M.~A., Purohit, S., Prashanth, U.~S., Raff, E., Skowron, A.,
  Sutawika, L., and van~der Wal, O. (2023).
\newblock Pythia: A suite for analyzing large language models across training
  and scaling.

\bibitem[Bills et~al., 2023]{bills2023language}
Bills, S., Cammarata, N., Mossing, D., Tillman, H., Gao, L., Goh, G.,
  Sutskever, I., Leike, J., Wu, J., and Saunders, W. (2023).
\newblock Language models can explain neurons in language models.
\newblock
  \url{https://openaipublic.blob.core.windows.net/neuron-explainer/paper/index.html}.

\bibitem[Boix-Adsera et~al., 2022]{boix-adsera_gulp_2022}
Boix-Adsera, E., Lawrence, H., Stepaniants, G., and Rigollet, P. (2022).
\newblock {GULP}: a prediction-based metric between representations.

\bibitem[Bommasani et~al., 2021]{bommasani2021opportunities}
Bommasani, R., Hudson, D.~A., Adeli, E., Altman, R., Arora, S., von Arx, S.,
  Bernstein, M.~S., Bohg, J., Bosselut, A., Brunskill, E., et~al. (2021).
\newblock On the opportunities and risks of foundation models.
\newblock {\em arXiv preprint arXiv:2108.07258}.

\bibitem[Bricken et~al., 2023]{bricken2023monosemanticity}
Bricken, T., Templeton, A., Batson, J., Chen, B., Jermyn, A., Conerly, T.,
  Turner, N., Anil, C., Denison, C., Askell, A., Lasenby, R., Wu, Y., Kravec,
  S., Schiefer, N., Maxwell, T., Joseph, N., Hatfield-Dodds, Z., Tamkin, A.,
  Nguyen, K., McLean, B., Burke, J.~E., Hume, T., Carter, S., Henighan, T., and
  Olah, C. (2023).
\newblock Towards monosemanticity: Decomposing language models with dictionary
  learning.
\newblock {\em Transformer Circuits Thread}.
\newblock
  https://transformer-circuits.pub/2023/monosemantic-features/index.html.

\bibitem[Brody et~al., 2023]{brody2023expressivity}
Brody, S., Alon, U., and Yahav, E. (2023).
\newblock On the expressivity role of layernorm in transformers' attention.
\newblock {\em arXiv preprint arXiv:2305.02582}.

\bibitem[Brown et~al., 2023]{brown2023privileged}
Brown, D., Vyas, N., and Bansal, Y. (2023).
\newblock On privileged and convergent bases in neural network representations.
\newblock {\em arXiv preprint arXiv:2307.12941}.

\bibitem[Bubeck et~al., 2023]{bubeck2023sparks}
Bubeck, S., Chandrasekaran, V., Eldan, R., Gehrke, J., Horvitz, E., Kamar, E.,
  Lee, P., Lee, Y.~T., Li, Y., Lundberg, S., et~al. (2023).
\newblock Sparks of artificial general intelligence: Early experiments with
  gpt-4.
\newblock {\em arXiv preprint arXiv:2303.12712}.

\bibitem[Burns et~al., 2022]{burns2022discovering}
Burns, C., Ye, H., Klein, D., and Steinhardt, J. (2022).
\newblock Discovering latent knowledge in language models without supervision.
\newblock {\em arXiv preprint arXiv:2212.03827}.

\bibitem[Cammarata et~al., 2021]{cammarata2021curve}
Cammarata, N., Goh, G., Carter, S., Voss, C., Schubert, L., and Olah, C.
  (2021).
\newblock Curve circuits.
\newblock {\em Distill}.
\newblock https://distill.pub/2020/circuits/curve-circuits.

\bibitem[Carlsmith, 2023]{carlsmith2023scheming}
Carlsmith, J. (2023).
\newblock Scheming ais: Will ais fake alignment during training in order to get
  power?
\newblock {\em arXiv preprint arXiv:2311.08379}.

\bibitem[Casper et~al., 2021]{casper2021frivolous}
Casper, S., Boix, X., D'Amario, V., Guo, L., Schrimpf, M., Vinken, K., and
  Kreiman, G. (2021).
\newblock Frivolous units: Wider networks are not really that wide.
\newblock In {\em Proceedings of the AAAI Conference on Artificial
  Intelligence}, volume~35, pages 6921--6929.

\bibitem[Chughtai et~al., 2023]{chughtai2023toy}
Chughtai, B., Chan, L., and Nanda, N. (2023).
\newblock A toy model of universality: Reverse engineering how networks learn
  group operations.
\newblock In {\em Proceedings of the 40th International Conference on Machine
  Learning}, ICML'23. JMLR.org.

\bibitem[Conmy et~al., 2023]{conmy2023towards}
Conmy, A., Mavor-Parker, A.~N., Lynch, A., Heimersheim, S., and Garriga-Alonso,
  A. (2023).
\newblock Towards automated circuit discovery for mechanistic interpretability.
\newblock {\em arXiv preprint arXiv:2304.14997}.

\bibitem[Cunningham et~al., 2023]{cunningham2023sparse}
Cunningham, H., Ewart, A., Riggs, L., Huben, R., and Sharkey, L. (2023).
\newblock Sparse autoencoders find highly interpretable features in language
  models.
\newblock {\em arXiv preprint arXiv:2309.08600}.

\bibitem[Dai et~al., 2021]{dai2021knowledge}
Dai, D., Dong, L., Hao, Y., Sui, Z., Chang, B., and Wei, F. (2021).
\newblock Knowledge neurons in pretrained transformers.
\newblock {\em arXiv preprint arXiv:2104.08696}.

\bibitem[Dalvi et~al., 2019]{dalvi2019one}
Dalvi, F., Durrani, N., Sajjad, H., Belinkov, Y., Bau, A., and Glass, J.
  (2019).
\newblock What is one grain of sand in the desert? analyzing individual neurons
  in deep nlp models.
\newblock In {\em Proceedings of the AAAI Conference on Artificial
  Intelligence}, volume~33, pages 6309--6317.

\bibitem[Dalvi et~al., 2020]{dalvi2020analyzing}
Dalvi, F., Sajjad, H., Durrani, N., and Belinkov, Y. (2020).
\newblock Analyzing redundancy in pretrained transformer models.
\newblock {\em arXiv preprint arXiv:2004.04010}.

\bibitem[Dar et~al., 2022]{dar2022analyzing}
Dar, G., Geva, M., Gupta, A., and Berant, J. (2022).
\newblock Analyzing transformers in embedding space.
\newblock {\em arXiv preprint arXiv:2209.02535}.

\bibitem[Ding et~al., 2021]{ding_grounding_2021}
Ding, F., Denain, J.-S., and Steinhardt, J. (2021).
\newblock Grounding {Representation} {Similarity} with {Statistical} {Testing}.

\bibitem[Donnelly and Roegiest, 2019]{donnelly2019interpretability}
Donnelly, J. and Roegiest, A. (2019).
\newblock On interpretability and feature representations: an analysis of the
  sentiment neuron.
\newblock In {\em Advances in Information Retrieval: 41st European Conference
  on IR Research, ECIR 2019, Cologne, Germany, April 14--18, 2019, Proceedings,
  Part I 41}, pages 795--802. Springer.

\bibitem[Doshi-Velez and Kim, 2017]{doshi2017towards}
Doshi-Velez, F. and Kim, B. (2017).
\newblock Towards a rigorous science of interpretable machine learning.
\newblock {\em arXiv preprint arXiv:1702.08608}.

\bibitem[Dravid et~al., 2023]{dravid2023rosetta}
Dravid, A., Gandelsman, Y., Efros, A.~A., and Shocher, A. (2023).
\newblock Rosetta neurons: Mining the common units in a model zoo.
\newblock In {\em Proceedings of the IEEE/CVF International Conference on
  Computer Vision}, pages 1934--1943.

\bibitem[Duong et~al., 2023]{duong_representational_2023}
Duong, L.~R., Zhou, J., Nassar, J., Berman, J., Olieslagers, J., and Williams,
  A.~H. (2023).
\newblock Representational dissimilarity metric spaces for stochastic neural
  networks.

\bibitem[Durrani et~al., 2022]{durrani2022linguistic}
Durrani, N., Dalvi, F., and Sajjad, H. (2022).
\newblock Linguistic correlation analysis: Discovering salient neurons in
  deepnlp models.
\newblock {\em arXiv preprint arXiv:2206.13288}.

\bibitem[Elhage et~al., 2022a]{elhage2022solu}
Elhage, N., Hume, T., Olsson, C., Nanda, N., Henighan, T., Johnston, S.,
  ElShowk, S., Joseph, N., DasSarma, N., Mann, B., Hernandez, D., Askell, A.,
  Ndousse, K., Drain, D., Chen, A., Bai, Y., Ganguli, D., Lovitt, L.,
  Hatfield-Dodds, Z., Kernion, J., Conerly, T., Kravec, S., Fort, S., Kadavath,
  S., Jacobson, J., Tran-Johnson, E., Kaplan, J., Clark, J., Brown, T.,
  McCandlish, S., Amodei, D., and Olah, C. (2022a).
\newblock Softmax linear units.
\newblock {\em Transformer Circuits Thread}.
\newblock https://transformer-circuits.pub/2022/solu/index.html.

\bibitem[Elhage et~al., 2022b]{elhage2022toy}
Elhage, N., Hume, T., Olsson, C., Schiefer, N., Henighan, T., Kravec, S.,
  Hatfield-Dodds, Z., Lasenby, R., Drain, D., Chen, C., et~al. (2022b).
\newblock Toy models of superposition.
\newblock {\em arXiv preprint arXiv:2209.10652}.

\bibitem[Elhage et~al., 2021]{elhage2021mathematical}
Elhage, N., Nanda, N., Olsson, C., Henighan, T., Joseph, N., Mann, B., Askell,
  A., Bai, Y., Chen, A., Conerly, T., et~al. (2021).
\newblock A mathematical framework for transformer circuits.
\newblock {\em Transformer Circuits Thread}.

\bibitem[Feng and Steinhardt, 2023]{feng2023language}
Feng, J. and Steinhardt, J. (2023).
\newblock How do language models bind entities in context?
\newblock {\em arXiv preprint arXiv:2310.17191}.

\bibitem[Gao et~al., 2020]{gao2020pile}
Gao, L., Biderman, S., Black, S., Golding, L., Hoppe, T., Foster, C., Phang,
  J., He, H., Thite, A., Nabeshima, N., et~al. (2020).
\newblock The pile: An 800gb dataset of diverse text for language modeling.
\newblock {\em arXiv preprint arXiv:2101.00027}.

\bibitem[Geva et~al., 2022]{geva2022transformer}
Geva, M., Caciularu, A., Wang, K.~R., and Goldberg, Y. (2022).
\newblock Transformer feed-forward layers build predictions by promoting
  concepts in the vocabulary space.
\newblock {\em arXiv preprint arXiv:2203.14680}.

\bibitem[Geva et~al., 2020]{geva2020transformer}
Geva, M., Schuster, R., Berant, J., and Levy, O. (2020).
\newblock Transformer feed-forward layers are key-value memories.
\newblock {\em arXiv preprint arXiv:2012.14913}.

\bibitem[Godfrey et~al., 2023]{godfrey_symmetries_2023}
Godfrey, C., Brown, D., Emerson, T., and Kvinge, H. (2023).
\newblock On the {Symmetries} of {Deep} {Learning} {Models} and their
  {Internal} {Representations}.

\bibitem[Goh et~al., 2021]{goh2021multimodal}
Goh, G., Cammarata, N., Voss, C., Carter, S., Petrov, M., Schubert, L.,
  Radford, A., and Olah, C. (2021).
\newblock Multimodal neurons in artificial neural networks.
\newblock {\em Distill}, 6(3):e30.

\bibitem[Gould et~al., 2023]{gould2023successor}
Gould, R., Ong, E., Ogden, G., and Conmy, A. (2023).
\newblock Successor heads: Recurring, interpretable attention heads in the
  wild.
\newblock {\em arXiv preprint arXiv:2312.09230}.

\bibitem[Gurnee et~al., 2023]{gurnee2023finding}
Gurnee, W., Nanda, N., Pauly, M., Harvey, K., Troitskii, D., and Bertsimas, D.
  (2023).
\newblock Finding neurons in a haystack: Case studies with sparse probing.
\newblock {\em arXiv preprint arXiv:2305.01610}.

\bibitem[Gurnee and Tegmark, 2023]{gurnee2023language}
Gurnee, W. and Tegmark, M. (2023).
\newblock Language models represent space and time.
\newblock {\em arXiv preprint arXiv:2310.02207}.

\bibitem[Gwilliam and Shrivastava, 2022]{gwilliam_beyond_2022}
Gwilliam, M. and Shrivastava, A. (2022).
\newblock Beyond {Supervised} vs. {Unsupervised}: {Representative}
  {Benchmarking} and {Analysis} of {Image} {Representation} {Learning}.

\bibitem[Hamilton et~al., 2018]{hamilton_diachronic_2018}
Hamilton, W.~L., Leskovec, J., and Jurafsky, D. (2018).
\newblock Diachronic {Word} {Embeddings} {Reveal} {Statistical} {Laws} of
  {Semantic} {Change}.

\bibitem[Hamrick and Mohamed, 2020]{hamrick2020levels}
Hamrick, J. and Mohamed, S. (2020).
\newblock Levels of analysis for machine learning.
\newblock {\em arXiv preprint arXiv:2004.05107}.

\bibitem[Hendel et~al., 2023]{hendel2023context}
Hendel, R., Geva, M., and Globerson, A. (2023).
\newblock In-context learning creates task vectors.
\newblock {\em arXiv preprint arXiv:2310.15916}.

\bibitem[Hendrycks and Gimpel, 2016]{hendrycks2016gaussian}
Hendrycks, D. and Gimpel, K. (2016).
\newblock Gaussian error linear units (gelus).
\newblock {\em arXiv preprint arXiv:1606.08415}.

\bibitem[Hendrycks et~al., 2023]{hendrycks2023overview}
Hendrycks, D., Mazeika, M., and Woodside, T. (2023).
\newblock An overview of catastrophic ai risks.
\newblock {\em arXiv preprint arXiv:2306.12001}.

\bibitem[Honnibal et~al., 2020]{honnibal2020spacy}
Honnibal, M., Montani, I., Van~Landeghem, S., and Boyd, A. (2020).
\newblock spacy: Industrial-strength natural language processing in python.

\bibitem[Hryniowski and Wong, 2020]{hryniowski_inter-layer_2020}
Hryniowski, A. and Wong, A. (2020).
\newblock Inter-layer {Information} {Similarity} {Assessment} of {Deep}
  {Neural} {Networks} {Via} {Topological} {Similarity} and {Persistence}
  {Analysis} of {Data} {Neighbour} {Dynamics}.

\bibitem[Huang et~al., 2023]{huang2023rigorously}
Huang, J., Geiger, A., D'Oosterlinck, K., Wu, Z., and Potts, C. (2023).
\newblock Rigorously assessing natural language explanations of neurons.
\newblock {\em arXiv preprint arXiv:2309.10312}.

\bibitem[Jastrz{\k{e}}bski et~al., 2017]{jastrzkebski2017residual}
Jastrz{\k{e}}bski, S., Arpit, D., Ballas, N., Verma, V., Che, T., and Bengio,
  Y. (2017).
\newblock Residual connections encourage iterative inference.
\newblock {\em arXiv preprint arXiv:1710.04773}.

\bibitem[Karamcheti et~al., 2021]{Mistral}
Karamcheti, S., Orr, L., Bolton, J., Zhang, T., Goel, K., Narayan, A.,
  Bommasani, R., Narayanan, D., Hashimoto, T., Jurafsky, D., Manning, C.~D.,
  Potts, C., Ré, C., and Liang, P. (2021).
\newblock Mistral - a journey towards reproducible language model training.

\bibitem[Khrulkov and Oseledets, 2018]{khrulkov_geometry_2018}
Khrulkov, V. and Oseledets, I. (2018).
\newblock Geometry {Score}: {A} {Method} {For} {Comparing} {Generative}
  {Adversarial} {Networks}.

\bibitem[Klabunde et~al., 2023]{klabunde_similarity_2023}
Klabunde, M., Schumacher, T., Strohmaier, M., and Lemmerich, F. (2023).
\newblock Similarity of {Neural} {Network} {Models}: {A} {Survey} of
  {Functional} and {Representational} {Measures}.

\bibitem[Kornblith et~al., 2019]{kornblith2019similarity}
Kornblith, S., Norouzi, M., Lee, H., and Hinton, G. (2019).
\newblock Similarity of neural network representations revisited.
\newblock In {\em International conference on machine learning}, pages
  3519--3529. PMLR.

\bibitem[Lange et~al., 2022]{lange_clustering_2022}
Lange, R.~D., Rolnick, D.~S., and Kording, K.~P. (2022).
\newblock Clustering units in neural networks: upstream vs downstream
  information.

\bibitem[Li et~al., 2015]{li2015convergent}
Li, Y., Yosinski, J., Clune, J., Lipson, H., and Hopcroft, J. (2015).
\newblock Convergent learning: Do different neural networks learn the same
  representations?
\newblock {\em arXiv preprint arXiv:1511.07543}.

\bibitem[Li et~al., 2016]{li_convergent_2016}
Li, Y., Yosinski, J., Clune, J., Lipson, H., and Hopcroft, J. (2016).
\newblock Convergent {Learning}: {Do} different neural networks learn the same
  representations?

\bibitem[Liao et~al., 2023]{liao2023generating}
Liao, I., Liu, Z., and Tegmark, M. (2023).
\newblock Generating interpretable networks using hypernetworks.
\newblock {\em arXiv preprint arXiv:2312.03051}.

\bibitem[Lim and Lauw, 2023]{lim2023disentangling}
Lim, J. and Lauw, H. (2023).
\newblock Disentangling transformer language models as superposed topic models.
\newblock In {\em Proceedings of the 2023 Conference on Empirical Methods in
  Natural Language Processing}, pages 8646--8666.

\bibitem[Lin, 2022]{lin_geometric_2022}
Lin, B. (2022).
\newblock Geometric and {Topological} {Inference} for {Deep} {Representations}
  of {Complex} {Networks}.
\newblock In {\em Companion {Proceedings} of the {Web} {Conference} 2022},
  pages 334--338.

\bibitem[Lu et~al., 2022]{lu_understanding_2022}
Lu, Y., Yang, W., Zhang, Y., Chen, Z., Chen, J., Xuan, Q., Wang, Z., and Yang,
  X. (2022).
\newblock Understanding the {Dynamics} of {DNNs} {Using} {Graph} {Modularity}.

\bibitem[Marr, 2010]{marr2010vision}
Marr, D. (2010).
\newblock {\em Vision: A computational investigation into the human
  representation and processing of visual information}.
\newblock MIT press.

\bibitem[McDougall et~al., 2023]{mcdougall2023copy}
McDougall, C., Conmy, A., Rushing, C., McGrath, T., and Nanda, N. (2023).
\newblock Copy suppression: Comprehensively understanding an attention head.
\newblock {\em arXiv preprint arXiv:2310.04625}.

\bibitem[McGrath et~al., 2023]{mcgrath2023hydra}
McGrath, T., Rahtz, M., Kramar, J., Mikulik, V., and Legg, S. (2023).
\newblock The hydra effect: Emergent self-repair in language model
  computations.
\newblock {\em arXiv preprint arXiv:2307.15771}.

\bibitem[Merullo et~al., 2023]{merullo2023circuit}
Merullo, J., Eickhoff, C., and Pavlick, E. (2023).
\newblock Circuit component reuse across tasks in transformer language models.
\newblock {\em arXiv preprint arXiv:2310.08744}.

\bibitem[Morcos et~al., 2018]{morcos_insights_2018}
Morcos, A.~S., Raghu, M., and Bengio, S. (2018).
\newblock Insights on representational similarity in neural networks with
  canonical correlation.

\bibitem[Mu and Andreas, 2020]{mu2020compositional}
Mu, J. and Andreas, J. (2020).
\newblock Compositional explanations of neurons.
\newblock {\em Advances in Neural Information Processing Systems},
  33:17153--17163.

\bibitem[Nanda, 2022]{nandatransformerlens2022}
Nanda, N. (2022).
\newblock Transformerlens.

\bibitem[Nanda et~al., 2023]{nanda2023progress}
Nanda, N., Chan, L., Liberum, T., Smith, J., and Steinhardt, J. (2023).
\newblock Progress measures for grokking via mechanistic interpretability.
\newblock {\em arXiv preprint arXiv:2301.05217}.

\bibitem[Ngo et~al., 2023]{ngo2023alignment}
Ngo, R., Chan, L., and Mindermann, S. (2023).
\newblock The alignment problem from a deep learning perspective.

\bibitem[Nguyen et~al., 2016]{nguyen2016multifaceted}
Nguyen, A., Yosinski, J., and Clune, J. (2016).
\newblock Multifaceted feature visualization: Uncovering the different types of
  features learned by each neuron in deep neural networks.

\bibitem[Nostalgebraist, 2020]{nostalgebraist2020interpreting}
Nostalgebraist (2020).
\newblock Interpreting gpt: The logit lens.
\newblock
  \url{https://www.alignmentforum.org/posts/AcKRB8wDpdaN6v6ru/interpreting-gpt-the-logit-lens}.

\bibitem[Olah, 2021]{olah2021interpretability}
Olah, C. (2021).
\newblock Interpretability vs neuroscience.
\newblock \url{https://colah.github.io/notes/interp-v-neuro/}.

\bibitem[Olah et~al., 2020a]{olah2020an}
Olah, C., Cammarata, N., Schubert, L., Goh, G., Petrov, M., and Carter, S.
  (2020a).
\newblock An overview of early vision in inceptionv1.
\newblock {\em Distill}.
\newblock https://distill.pub/2020/circuits/early-vision.

\bibitem[Olah et~al., 2020b]{olah2020zoom}
Olah, C., Cammarata, N., Schubert, L., Goh, G., Petrov, M., and Carter, S.
  (2020b).
\newblock Zoom in: An introduction to circuits.
\newblock {\em Distill}, 5(3):e00024--001.

\bibitem[Olsson et~al., 2022]{olsson2022context}
Olsson, C., Elhage, N., Nanda, N., Joseph, N., DasSarma, N., Henighan, T.,
  Mann, B., Askell, A., Bai, Y., Chen, A., Conerly, T., Drain, D., Ganguli, D.,
  Hatfield-Dodds, Z., Hernandez, D., Johnston, S., Jones, A., Kernion, J.,
  Lovitt, L., Ndousse, K., Amodei, D., Brown, T., Clark, J., Kaplan, J.,
  McCandlish, S., and Olah, C. (2022).
\newblock In-context learning and induction heads.
\newblock {\em Transformer Circuits Thread}.
\newblock
  https://transformer-circuits.pub/2022/in-context-learning-and-induction-heads/index.html.

\bibitem[Quirke et~al., 2023]{quirke2023training}
Quirke, L., Heindrich, L., Gurnee, W., and Nanda, N. (2023).
\newblock Training dynamics of contextual n-grams in language models.
\newblock {\em arXiv preprint arXiv:2311.00863}.

\bibitem[Radford et~al., 2017]{radford2017learning}
Radford, A., Jozefowicz, R., and Sutskever, I. (2017).
\newblock Learning to generate reviews and discovering sentiment.
\newblock {\em arXiv preprint arXiv:1704.01444}.

\bibitem[Radford et~al., 2018]{radford2018improving}
Radford, A., Narasimhan, K., Salimans, T., Sutskever, I., et~al. (2018).
\newblock Improving language understanding by generative pre-training.

\bibitem[Radford et~al., 2019]{radford2019language}
Radford, A., Wu, J., Child, R., Luan, D., Amodei, D., Sutskever, I., et~al.
  (2019).
\newblock Language models are unsupervised multitask learners.
\newblock {\em OpenAI blog}, 1(8):9.

\bibitem[Raghu et~al., 2017]{raghu_svcca_2017}
Raghu, M., Gilmer, J., Yosinski, J., and Sohl-Dickstein, J. (2017).
\newblock {SVCCA}: {Singular} {Vector} {Canonical} {Correlation} {Analysis} for
  {Deep} {Learning} {Dynamics} and {Interpretability}.

\bibitem[Reuther et~al., 2018]{reuther2018interactive}
Reuther, A., Kepner, J., Byun, C., Samsi, S., Arcand, W., Bestor, D., Bergeron,
  B., Gadepally, V., Houle, M., Hubbell, M., Jones, M., Klein, A., Milechin,
  L., Mullen, J., Prout, A., Rosa, A., Yee, C., and Michaleas, P. (2018).
\newblock Interactive supercomputing on 40,000 cores for machine learning and
  data analysis.
\newblock In {\em 2018 IEEE High Performance extreme Computing Conference
  (HPEC)}, pages 1--6. IEEE.

\bibitem[Sajjad et~al., 2022]{sajjad2022analyzing}
Sajjad, H., Durrani, N., Dalvi, F., Alam, F., Khan, A.~R., and Xu, J. (2022).
\newblock Analyzing encoded concepts in transformer language models.
\newblock {\em arXiv preprint arXiv:2206.13289}.

\bibitem[Schubert et~al., 2021a]{schubert2021high-low}
Schubert, L., Voss, C., Cammarata, N., Goh, G., and Olah, C. (2021a).
\newblock High-low frequency detectors.
\newblock {\em Distill}.
\newblock https://distill.pub/2020/circuits/frequency-edges.

\bibitem[Schubert et~al., 2021b]{schubert2021high}
Schubert, L., Voss, C., Cammarata, N., Goh, G., and Olah, C. (2021b).
\newblock High-low frequency detectors.
\newblock {\em Distill}, 6(1):e00024--005.

\bibitem[Shahbazi et~al., 2021]{shahbazi_using_2021}
Shahbazi, M., Shirali, A., Aghajan, H., and Nili, H. (2021).
\newblock Using distance on the {Riemannian} manifold to compare
  representations in brain and in models.
\newblock {\em NeuroImage}, 239:118271.

\bibitem[Tang et~al., 2020]{tang_similarity_2020}
Tang, S., Maddox, W.~J., Dickens, C., Diethe, T., and Damianou, A. (2020).
\newblock Similarity of {Neural} {Networks} with {Gradients}.

\bibitem[Todd et~al., 2023]{todd2023function}
Todd, E., Li, M.~L., Sharma, A.~S., Mueller, A., Wallace, B.~C., and Bau, D.
  (2023).
\newblock Function vectors in large language models.
\newblock {\em arXiv preprint arXiv:2310.15213}.

\bibitem[Variengien and Winsor, 2023]{variengien2023look}
Variengien, A. and Winsor, E. (2023).
\newblock Look before you leap: A universal emergent decomposition of retrieval
  tasks in language models.

\bibitem[Voita et~al., 2023]{voita2023neurons}
Voita, E., Ferrando, J., and Nalmpantis, C. (2023).
\newblock Neurons in large language models: Dead, n-gram, positional.
\newblock {\em arXiv preprint arXiv:2309.04827}.

\bibitem[Wang et~al., 2022a]{wang2022interpretability}
Wang, K., Variengien, A., Conmy, A., Shlegeris, B., and Steinhardt, J. (2022a).
\newblock Interpretability in the wild: a circuit for indirect object
  identification in gpt-2 small.
\newblock {\em arXiv preprint arXiv:2211.00593}.

\bibitem[Wang et~al., 2018]{wang_towards_2018}
Wang, L., Hu, L., Gu, J., Wu, Y., Hu, Z., He, K., and Hopcroft, J. (2018).
\newblock Towards {Understanding} {Learning} {Representations}: {To} {What}
  {Extent} {Do} {Different} {Neural} {Networks} {Learn} the {Same}
  {Representation}.

\bibitem[Wang and Isola, 2022]{wang_understanding_2022}
Wang, T. and Isola, P. (2022).
\newblock Understanding {Contrastive} {Representation} {Learning} through
  {Alignment} and {Uniformity} on the {Hypersphere}.

\bibitem[Wang et~al., 2022b]{wang2022finding}
Wang, X., Wen, K., Zhang, Z., Hou, L., Liu, Z., and Li, J. (2022b).
\newblock Finding skill neurons in pre-trained transformer-based language
  models.
\newblock {\em arXiv preprint arXiv:2211.07349}.

\bibitem[Weidinger et~al., 2022]{weidinger2022taxonomy}
Weidinger, L., Uesato, J., Rauh, M., Griffin, C., Huang, P.-S., Mellor, J.,
  Glaese, A., Cheng, M., Balle, B., Kasirzadeh, A., et~al. (2022).
\newblock Taxonomy of risks posed by language models.
\newblock In {\em Proceedings of the 2022 ACM Conference on Fairness,
  Accountability, and Transparency}, pages 214--229.

\bibitem[Williams et~al., 2022]{williams_generalized_2022}
Williams, A.~H., Kunz, E., Kornblith, S., and Linderman, S.~W. (2022).
\newblock Generalized {Shape} {Metrics} on {Neural} {Representations}.

\bibitem[Xia et~al., 2023]{xia2023sheared}
Xia, M., Gao, T., Zeng, Z., and Chen, D. (2023).
\newblock Sheared llama: Accelerating language model pre-training via
  structured pruning.
\newblock {\em arXiv preprint arXiv:2310.06694}.

\bibitem[Xiao et~al., 2023]{xiao2023efficient}
Xiao, G., Tian, Y., Chen, B., Han, S., and Lewis, M. (2023).
\newblock Efficient streaming language models with attention sinks.
\newblock {\em arXiv preprint arXiv:2309.17453}.

\bibitem[Xin et~al., 2019]{xin2019part}
Xin, J., Lin, J., and Yu, Y. (2019).
\newblock What part of the neural network does this? understanding lstms by
  measuring and dissecting neurons.
\newblock In {\em Proceedings of the 2019 Conference on Empirical Methods in
  Natural Language Processing and the 9th International Joint Conference on
  Natural Language Processing (EMNLP-IJCNLP)}, pages 5823--5830.

\bibitem[Zhang et~al., 2022]{zhang2022opt}
Zhang, S., Roller, S., Goyal, N., Artetxe, M., Chen, M., Chen, S., Dewan, C.,
  Diab, M., Li, X., Lin, X.~V., et~al. (2022).
\newblock Opt: Open pre-trained transformer language models.
\newblock {\em arXiv preprint arXiv:2205.01068}.

\bibitem[Zhong et~al., 2023]{zhong2023clock}
Zhong, Z., Liu, Z., Tegmark, M., and Andreas, J. (2023).
\newblock The clock and the pizza: Two stories in mechanistic explanation of
  neural networks.
\newblock {\em arXiv preprint arXiv:2306.17844}.

\bibitem[Zou et~al., 2023]{zou2023representation}
Zou, A., Phan, L., Chen, S., Campbell, J., Guo, P., Ren, R., Pan, A., Yin, X.,
  Mazeika, M., Dombrowski, A.-K., et~al. (2023).
\newblock Representation engineering: A top-down approach to ai transparency.
\newblock {\em arXiv preprint arXiv:2310.01405}.

\end{thebibliography}

\appendix

\section{Additional Empirical Details}

All of our code and data is available at \url{https://github.com/wesg52/universal-neurons}.

Most of our plots in the main text (and therefore neuron indices) correspond to the HuggingFace model \texttt{stanford-crfm/arwen-gpt2-medium-x21} with our additional correlation experiments being conducted on \texttt{stanford-crfm/alias-gpt2-small-x21} and \texttt{EleutherAI/pythia-160m}.
\subsection{Weight Preprocessing} \label{sec:weight_preprocessing}
We employ several standard weight preprocessing techniques to simplify calculations \citep{nandatransformerlens2022}.

\paragraph{Folding in Layer Norm}
% The calculation in \ref{eq:mlp} doesn't capture the whole story. This is because prior to every MLP calculation, a layer norm operation is applied to the residual stream before it is fed into the MLP. Luckily, the TransformerLens package is able to take care of this by folding the layer norm into the weights and biases of the MLP, creating matrices $W_\text{eff}$ and $b_\text{eff}$. The layer norm for some $x \in \R^n$ is 
Most layer norm implementations also include trainable parameters $\boldsymbol{\gamma} \in \R^n$ and $\mathbf{b} \in \R^n$
\begin{equation}
    \textnormal{LayerNorm($\mathbf{x}$)} = \frac{\mathbf{x}-\mathbb{E}(\mathbf{x})}{\sqrt{\text{Var}(\mathbf{x})}} * \boldsymbol{\gamma} + \mathbf{b}.
\end{equation}

To account for these, we can ``fold'' the layer norm parameters in to $W_\text{in}$ by observing that the layer norm parameters are equivalent to a linear layer, and then combine the adjacent linear layers. In particular, we can create effective weights
\begin{equation}
    \mathbf{W}_\text{eff} = \mathbf{W}_\text{in} ~ \mathbf{diag}(\boldsymbol{\gamma}) \quad  \quad \mathbf{b}_\text{eff} = \mathbf{b}_\text{in} + \mathbf{W}_\text{in} \mathbf{b}
\end{equation}
Finally, we can center the reading weights because the preceding layer norm projects out the all ones vector. Thus we can center the weights $\mathbf{W}_\text{eff}$ becomes \begin{equation*}
    \mathbf{W}_{\text{eff}}^{'}(i, :) = \mathbf{W}_{\text{eff}}(i, :) - \bar{\mathbf{W}}_{\text{eff}}(i, :)
\end{equation*}
% Accounting for the division by the square root of the variance is trickier because it is non-linear. Under the assumption that every feature contributes minimally to the total variance, we can approximate it as a constant at that position, and thus unimportant for our calculations. 
\paragraph{Writing Weight Centering}
Every time the model interacts with the residual stream it applies a LayerNorm first. Thus the components of $\mathbf{W}_\text{out}$ and $\mathbf{b}_\text{out}$ that lie along the all-ones direction of the residual stream have no effect on the model's calculation. So, we again mean-center $\mathbf{W}_\text{out}$ and $\mathbf{b}_\text{out}$ by subtracting the means of the columns of $\mathbf{W}_\text{out}$ 

\begin{equation*}
    \mathbf{W}_{\text{out}}^{'}(:, i) = \mathbf{W}_{\text{out}}(:, i) - \bar{\mathbf{W}}_{\text{out}}(:, i)
\end{equation*}

\paragraph{Unembed Centering} Additionally, since softmax is translation invariant, we modify $\mathbf{W}_U$ into 
\begin{equation*}
    \mathbf{W}_{\text{U}}^{'}(:, i) = \mathbf{W}_{\text{U}}(:, i) - \mathbb{\mathbf{w}}_i
\end{equation*}
For both of theses, see the transformer lens documentation for more details. 

The purpose of all of these translations is to remove irrelevant components and other parameterization degrees of freedom so that cosine similarities and other weight computations have mean 0.
\subsection{Correlation Computations} \label{sec:appendix_corr_computation}
We compute our correlations over a 100 million token subset of the Pile test set \citep{gao2020pile}, tokenized to a context length of 512 tokens. We compute correlations over all tokens that are not padding, beginning-of-sequence, or new-line tokens.

\paragraph{Efficient Computation}
Because storing neuron activations for two models over 100M tokens would be 36 petabytes of data, we require a streaming algorithm. To do so, observe that
given a pair of neuron activations $\left\{\left(x_1, y_1\right), \ldots,\left(x_n, y_n\right)\right\}$ consisting of  $n$ pairs, the correlation can be computed as 
$$
\rho_{x y}=\frac{\sum_{i=1}^n\left(x_i-\bar{x}\right)\left(y_i-\bar{y}\right)}{\sqrt{\sum_{i=1}^n\left(x_i-\bar{x}\right)^2} \sqrt{\sum_{i=1}^n\left(y_i-\bar{y}\right)^2}} =\frac{\sum_i x_i y_i-n \bar{x} \bar{y}}{\sqrt{\sum_i x_i^2-n \bar{x}^2} \sqrt{\sum_i y_i^2-n \bar{y}^2}}
$$
where $\bar{x}, \bar{y}$ are the sample mean.
Therefore, instead of saving all neuron activations, we can maintain four \texttt{n\_neuron} dimensional vectors and one $\texttt{n\_neuron} \times \texttt{n\_neuron}$ matrix corresponding to the running neuron activation means in model A and model B, a running sum of each neurons squared activation, and a running sum of pairwise products. At the end of the dataset, we perform the appropriate arithmetic to combine the results into pairwise correlations for all models.
% where $\bar{x}, \bar{y}$ are the sample mean.
% We store a few intermediate results after each batch of inference to prepare us for the computation of the final correlation results. We add up the two models' activation summing over all batched token position for each layer and neuron, with a shape of (\texttt{n\_layers}, \texttt{d\_mlp}), their corresponding element-wise squared matrices, and their layer-wise matrix product, with a shape of (\texttt{m1\_layers}, \texttt{m1\_d\_mlp}, \texttt{m2\_layers}, \texttt{m2\_d\_mlp}). With these intermediate results, we can compute the overall correlation layer-wise by computing the sum of the difference over the sum of the norms. 

\subsection{Model Hyperparameters} \label{sec:hparams}
\begin{table}[h]
\centering
\begin{tabular}{@{}lccc@{}} % @{} removes padding on th
 
\toprule % Top horizontal line
Property & GPT-2 Small & GPT-2 Medium & Pythia 160M \\
\midrule % Mid horizontal line (under headings)
layers & 12  & 24 & 12\\
heads & 12 & 16 & 12\\
$d_\text{model}$ & 768 & 1024 & 768\\
$d_\text{vocab}$  & 50257 & 50257 & 50304\\
$d_\text{MLP}$ & 3072 & 4096 &3072\\
parameters & 160M & 410M & 160M\\
context & 1024 & 1024 & 2048\\
activation function & gelu\_new & gelu\_new & gelu\\
pos embeddings & absolute & absolute & RoPE \\
rotary percentage & N/A & N/A & 25 \\
precision  & Float-32 & Float-32 & Float-16\\
dataset  & OpenwebText & OpenwebText & Pile\\
$p_\text{dropout}$  & 0.1 & 0.1 & 0\\
\bottomrule % Bottom horizontal line
\end{tabular}
\caption{Hyperparameters of models}
\end{table}

\newpage
\section{Additional Results}

\begin{figure}[h]
    \centering
    \includegraphics[width=\linewidth]{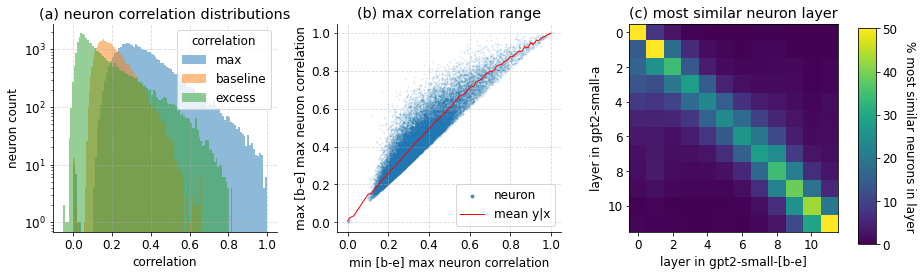}
    \caption{Summary of neuron correlation experiments in GPT2-small-a. (a) Distribution of the mean (over models b-e) max (over neurons) correlation, the mean baseline correlation, and the difference (excess). (b) The max (over models) max (over neurons) correlation compared to the min (over models) max (over neuron) correlation for each neuron. (c) Percentage of layer pairs with most similar neuron pairs.}
    \label{fig:corr_summary_small}
\end{figure}

\begin{figure}
    \centering
    \includegraphics[width=\linewidth]{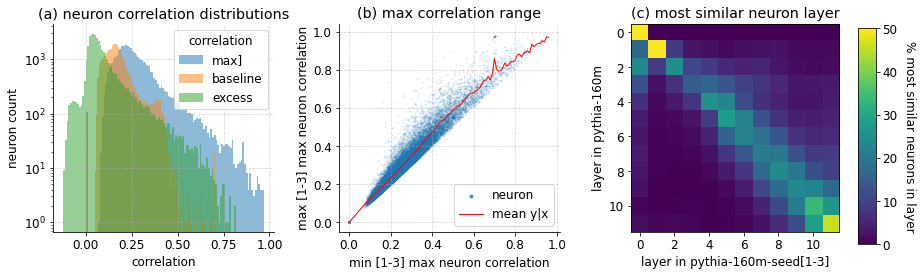}
    \caption{Summary of neuron correlation experiments in Pythia-160m. (a) Distribution of the mean (over models b-e) max (over neurons) correlation, the mean baseline correlation, and the difference (excess). (b) The max (over models) max (over neurons) correlation compared to the min (over models) max (over neuron) correlation for each neuron. (c) Percentage of layer pairs with most similar neuron pairs.}
    \label{fig:corr_summary_pythia}
\end{figure}

\begin{figure}
    \centering
    \vspace{-2em}
    \includegraphics[width=\linewidth]{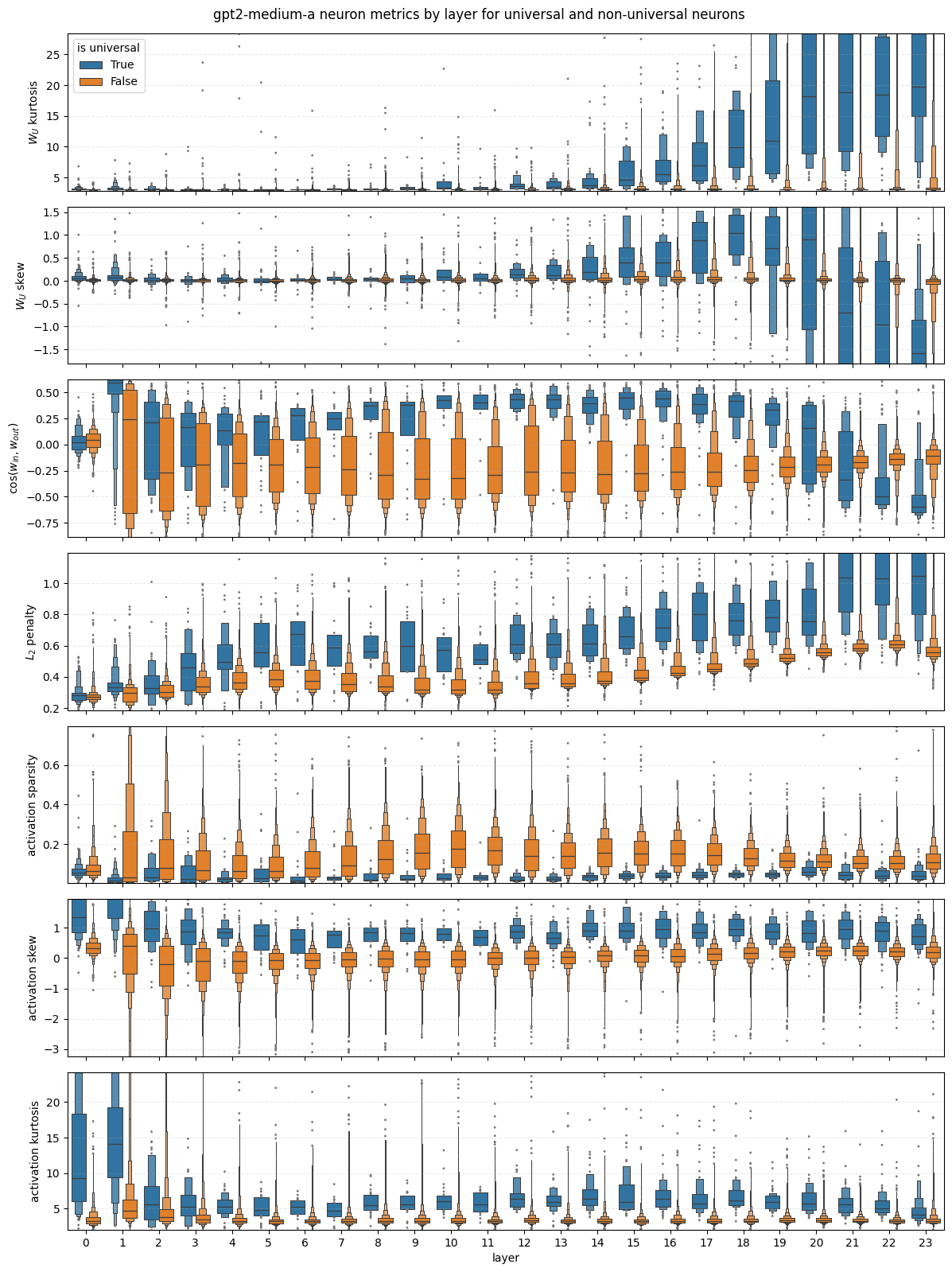}
    \caption{Distribution of neuron metrics for universal and non-universal neurons in GPT2-medium-a by layer. From top to bottom: the kurtosis of $\cos(\mathbf{W}_U, \mathbf{w}_{out})$, the skew of $\cos(\mathbf{W}_U, \mathbf{w}_{out})$, cosine similarity between input and output weight $\cos(\mathbf{w}_{in}, \mathbf{w}_{out})$, weight decay penalty $\|\mathbf{w}_{in}\|_2^2 + \|\mathbf{w}_{out}\|_2^2$, activation frequency (percentage of activations greater than 0), the pre-activation skew, and the pre-activation kurtosis.}
    \label{fig:properties_by_layer}
\end{figure}

\begin{figure}
    \centering
    \vspace{-3em}
    \includegraphics[width=\linewidth, height=52em]{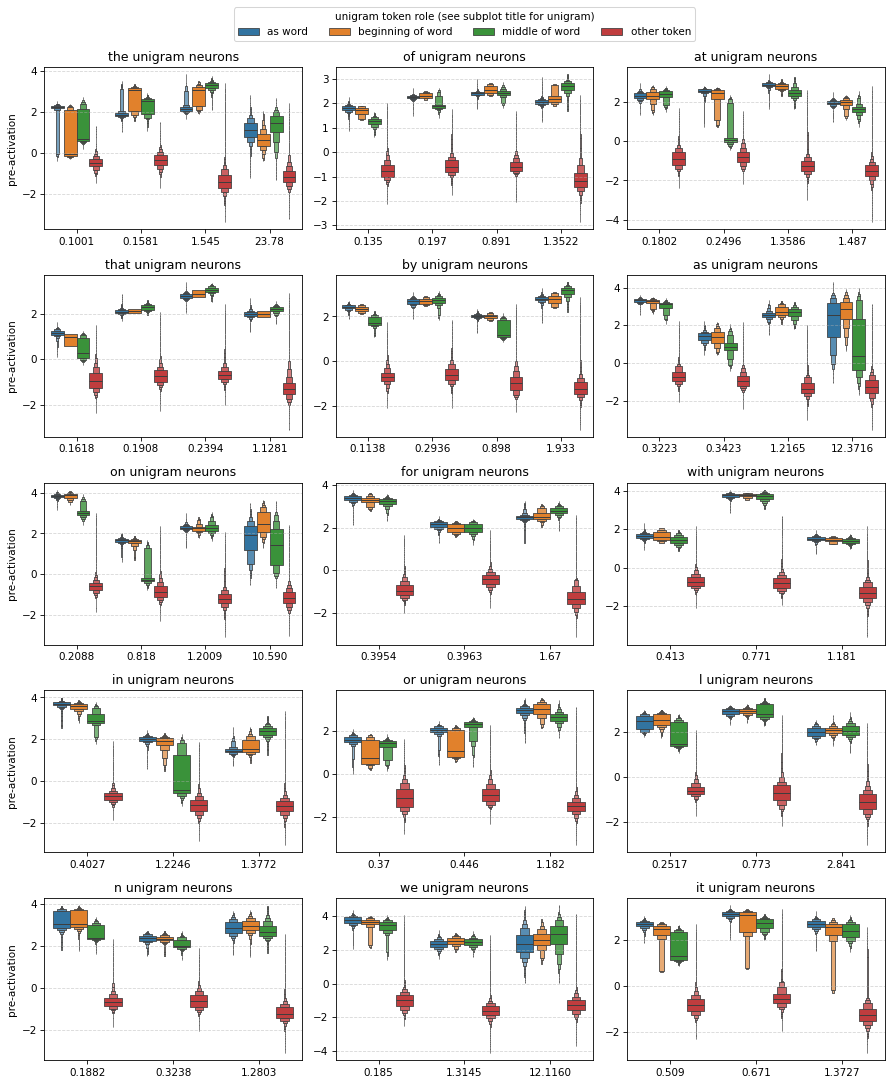}
    \includegraphics[width=\linewidth, height=11em]{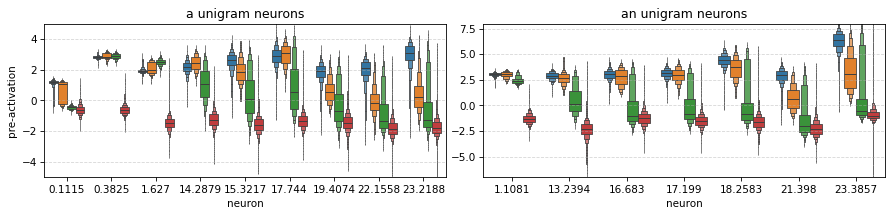}
    \caption{Duplicate unigram neurons in GPT2-medium-a. Each subplot depicts several neurons which activate on a particular token, broken down by whether this token exists as a standalone word, is the first token in a multi-token word, or is a non-first token in a multi-token word, versus all other tokens (e.g., ``an,'' ``an|agram,'' ``Gig|an|tism'').}
    \label{fig:unigram_duplicates}
\end{figure}

\begin{figure}
    \centering
    \includegraphics[width=\linewidth]{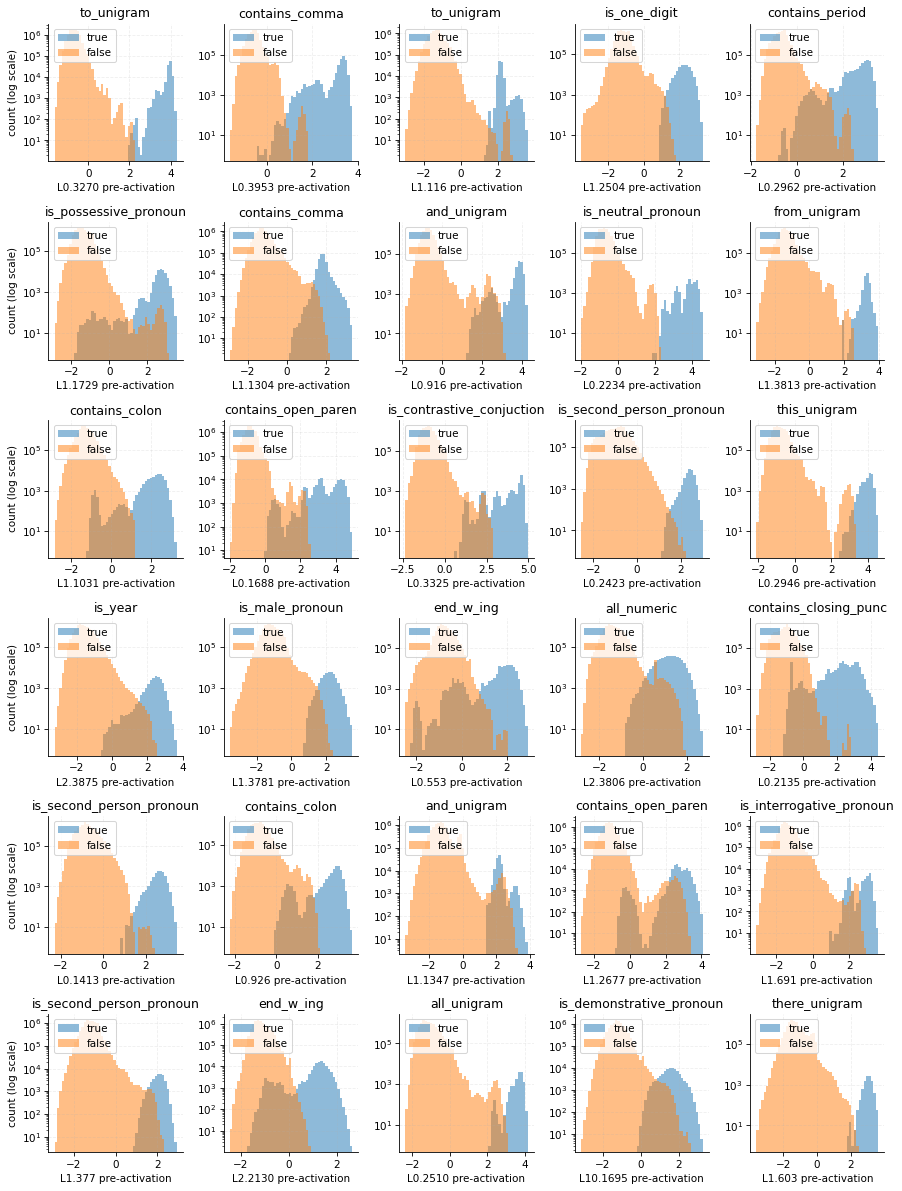}
    \caption{Universal unigram neurons in GPT2-medium-a.}
    \label{fig:fig:unigram_nonduplicate}
\end{figure}

\begin{figure}
    \centering
    \includegraphics[width=\linewidth]{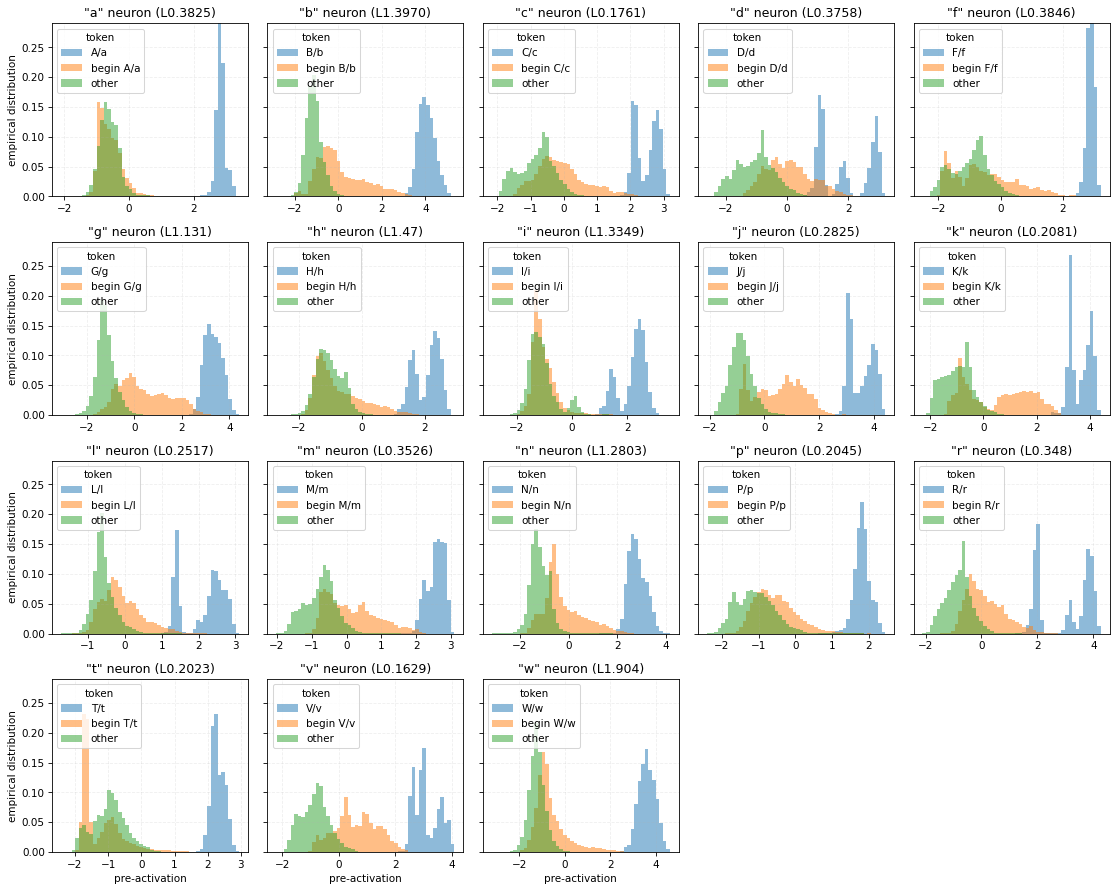}
    \caption{Universal alphabet neurons in GPT2-medium-a.}
    \label{fig:alphabet_neurons_full}
\end{figure}

\begin{figure}
    \centering
    \includegraphics[width=\linewidth]{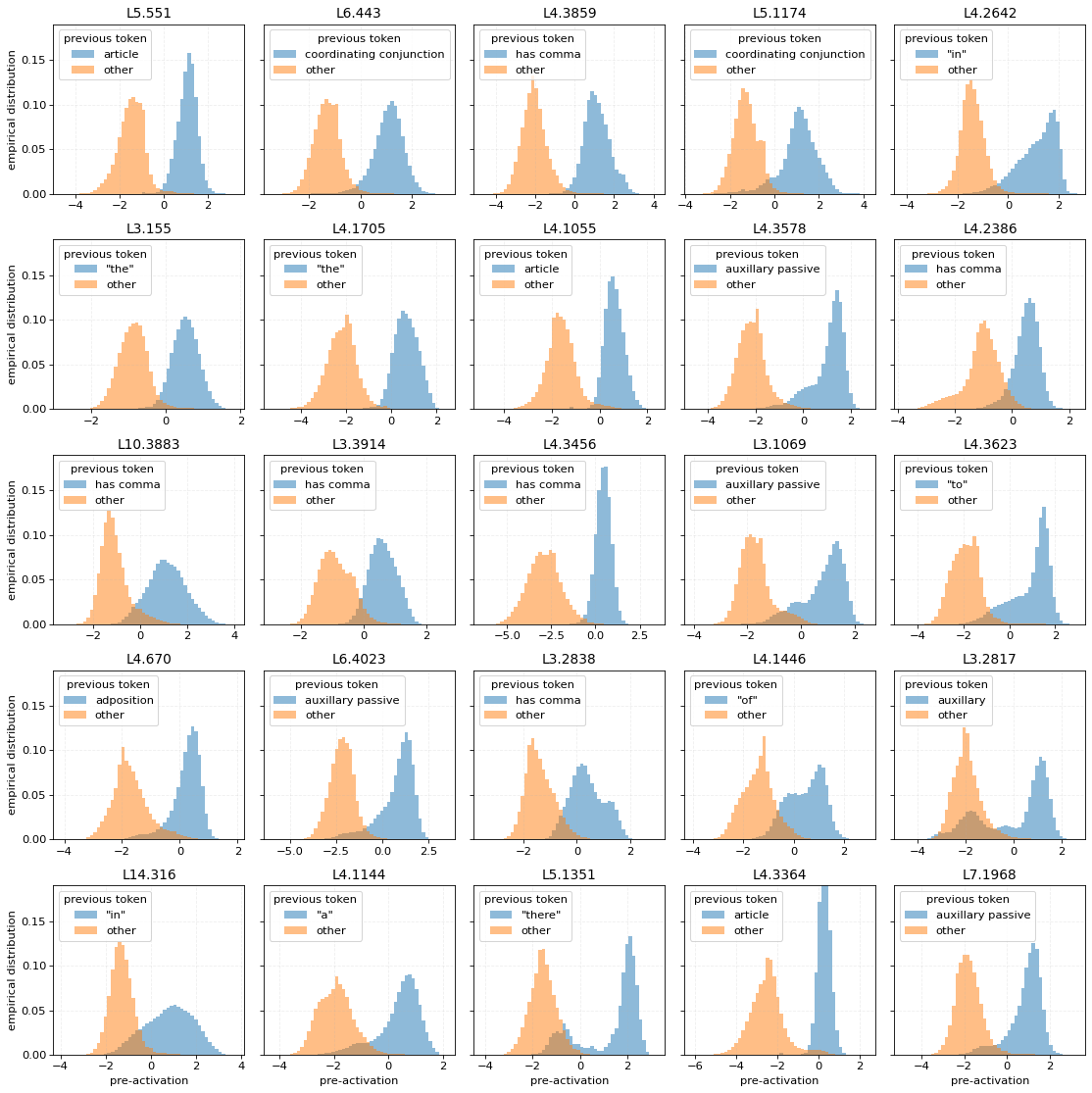}
    \caption{Universal previous token neurons in GPT2-medium-a.}
    \label{fig:prev_token_neurons_full}
\end{figure}

\begin{figure}
    \centering
    \includegraphics[width=\linewidth]{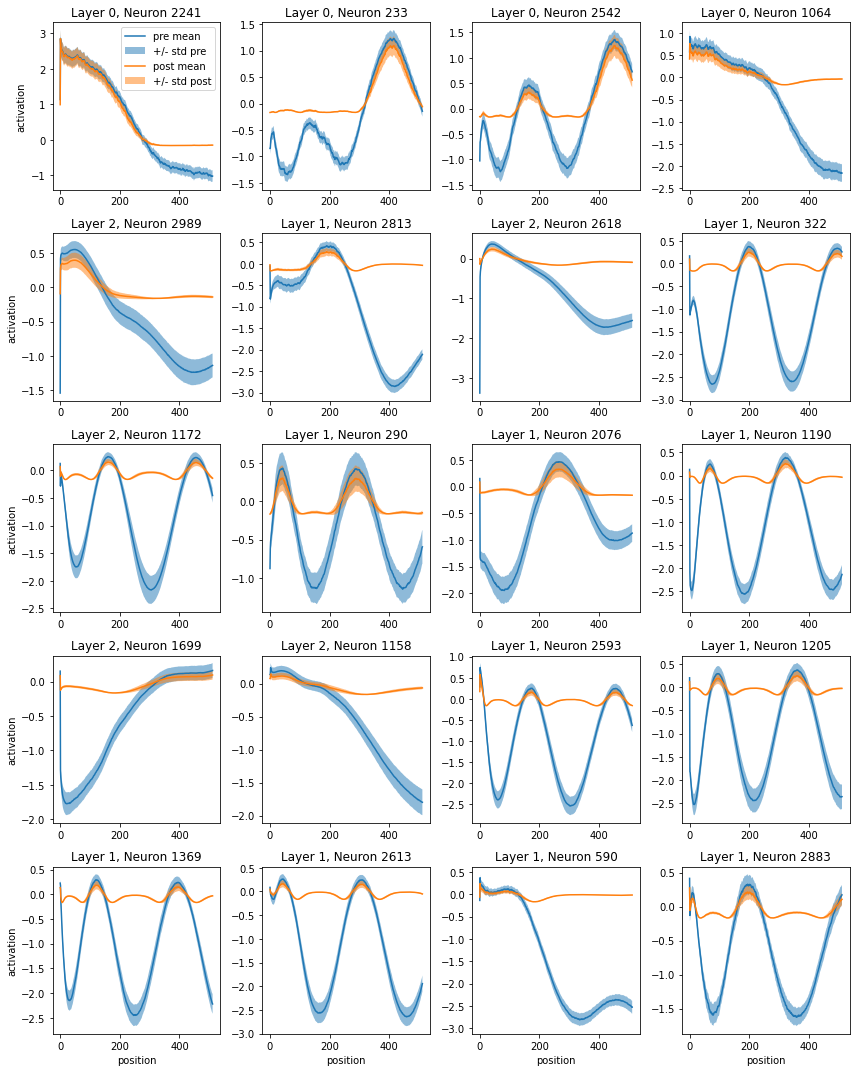}
    \caption{Universal position neurons in GPT2-small-a.}
    \label{fig:position_appendix}
\end{figure}

\begin{figure}
    \centering
    \includegraphics[width=\linewidth]{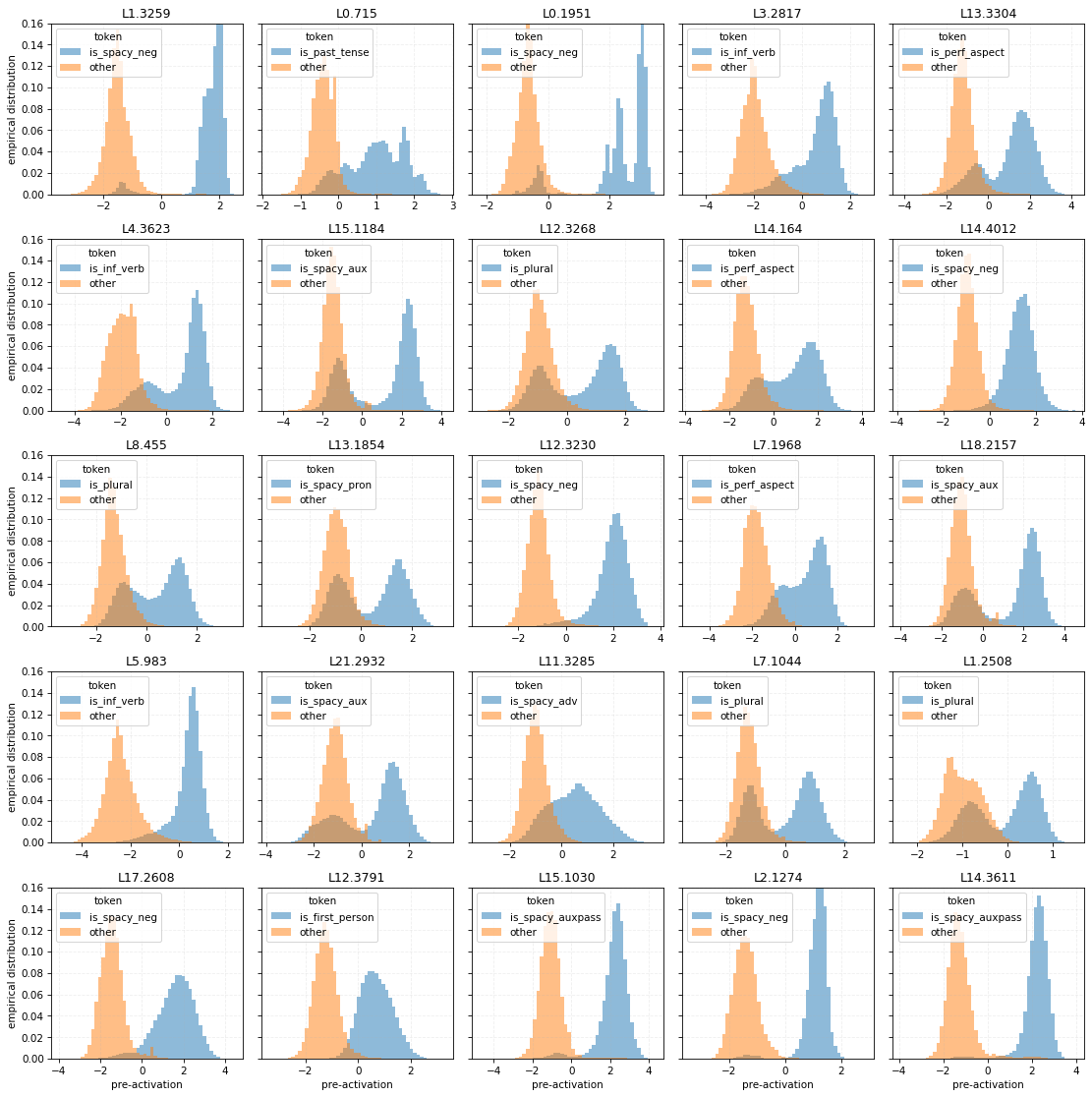}
    \caption{Universal syntax neurons in GPT2-medium-a.}
    \label{fig:syntax_neurons_full}
\end{figure}

\begin{figure}
    \centering
    \includegraphics[width=\linewidth]{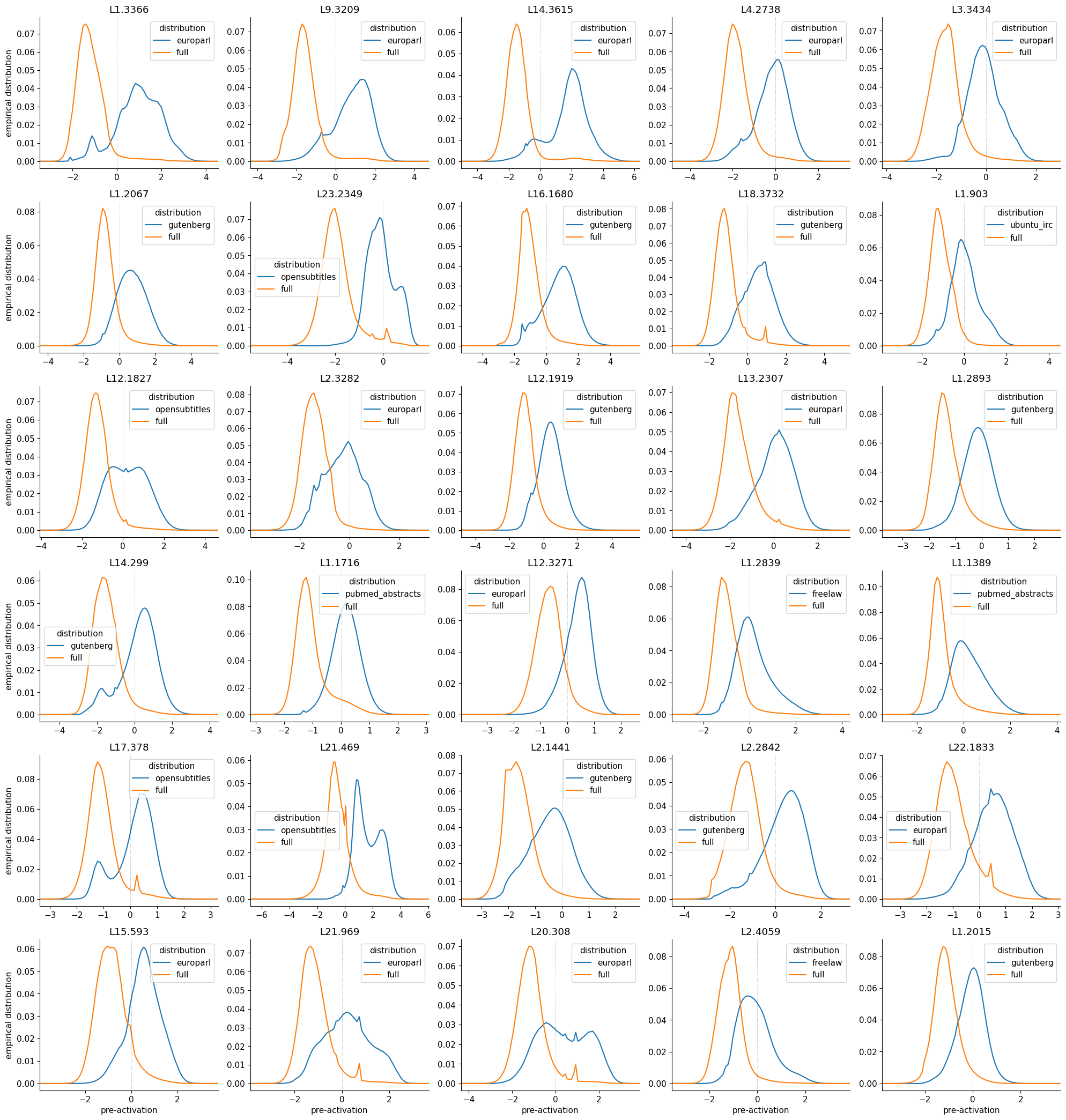}
    \caption{Universal context neurons in GPT2-medium-a.}
    \label{fig:context_appendix}
\end{figure}

\begin{figure}
    \centering
    \includegraphics[width=\linewidth]{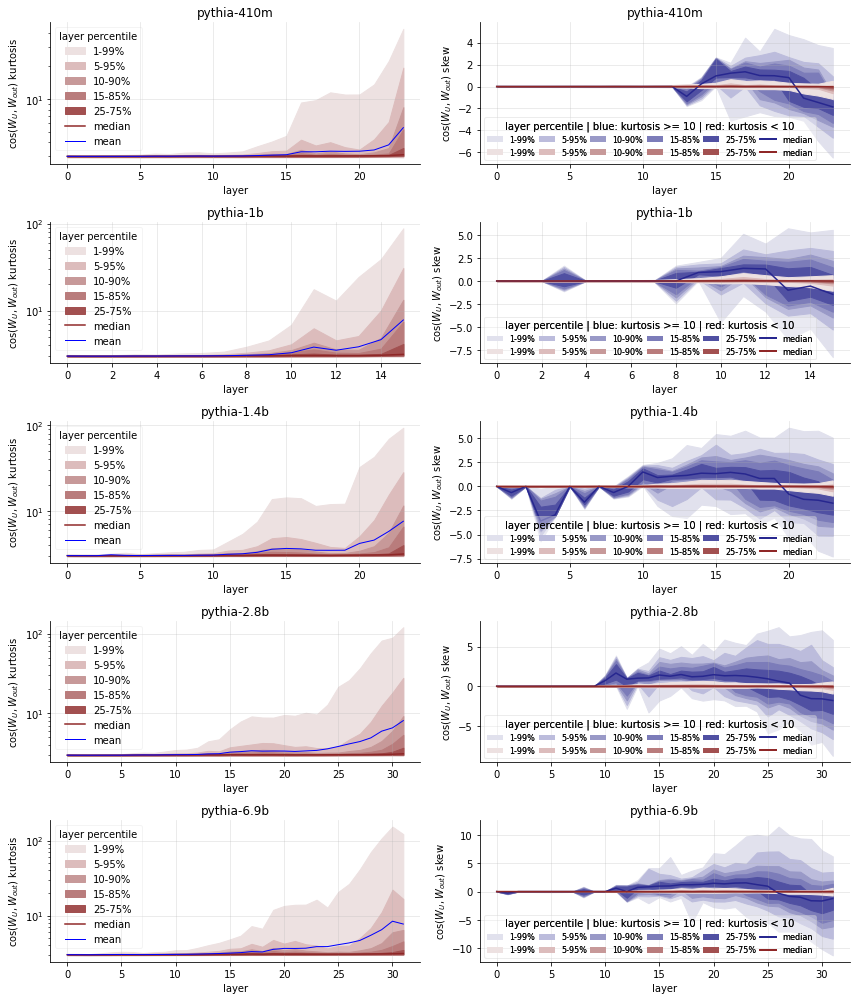}
    \caption{Distribution of vocabulary composition statistics for five different Pythia models measured over layers. Left shows percentiles of $\cos(\mathbf{W}_U, \mathbf{W}_{out})$ kurtosis. Right shows percentiles of $\cos(\mathbf{W}_U, \mathbf{W}_{out})$ skew broken down by whether neuron has $\cos(\mathbf{W}_U, \mathbf{W}_{out})$ kurtosis greater than or less than 10.}
    \label{fig:pythia_logit_stats}
\end{figure}

\begin{figure}
    \centering
    \includegraphics[width=\linewidth]{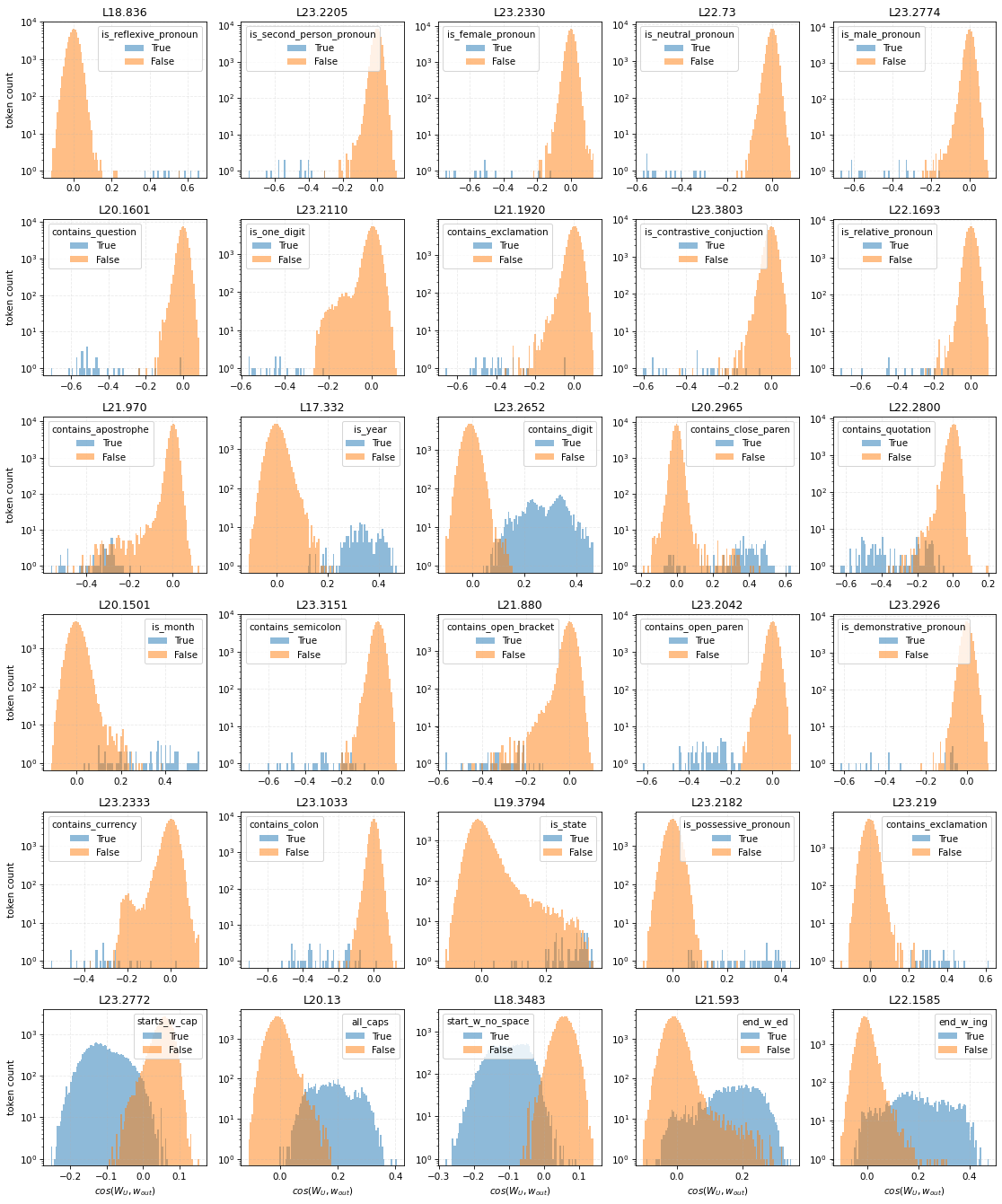}
    \caption{Universal prediction neurons in GPT2-medium-a.}
    \label{fig:prediction_neurons_full}
\end{figure}

\begin{figure}
    \centering
    \includegraphics[width=\linewidth]{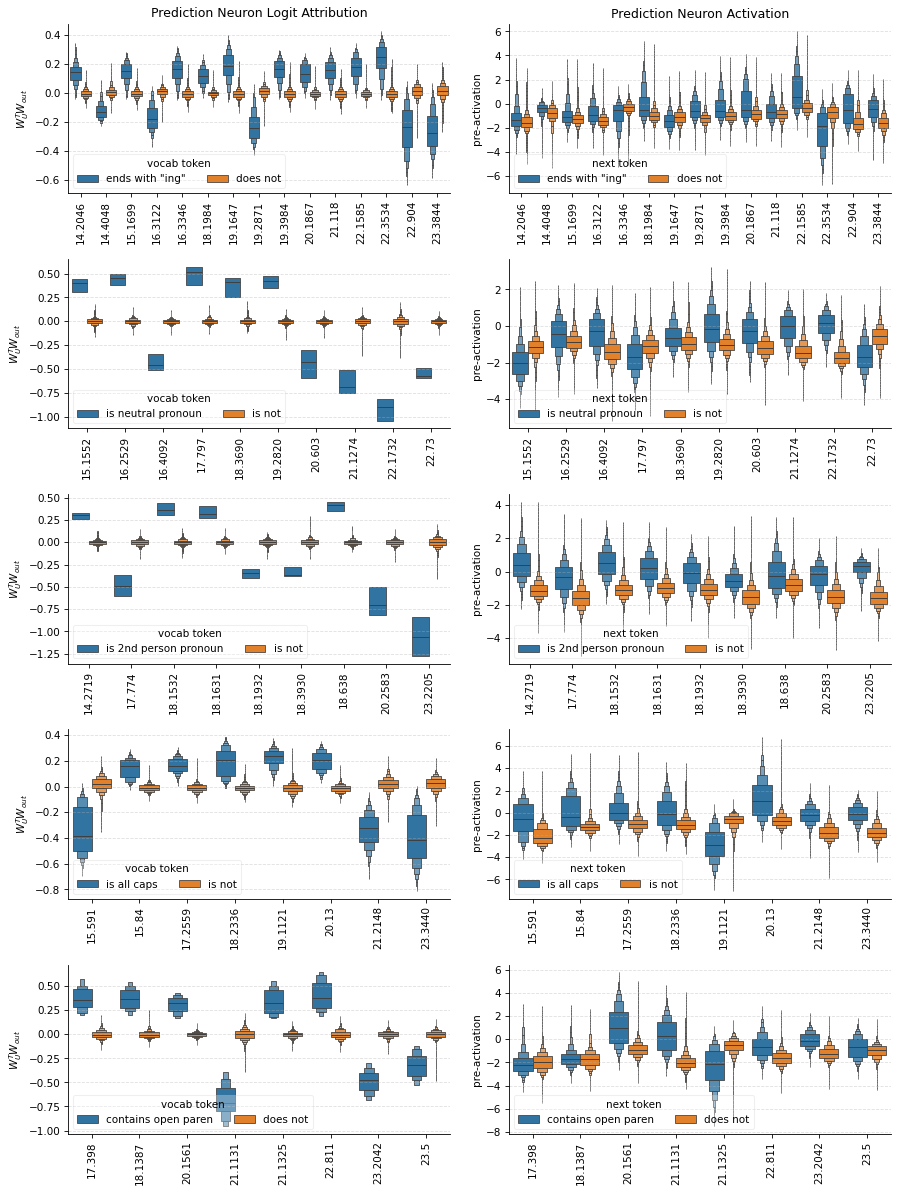}
    \caption{Prediction neurons for the same feature in GPT2-medium-a. Left column depicts logit effect broken down by vocabulary item per neuron and right column shows activation value broken down by true next token per neuron. }
    \label{fig:prediction_neurons_duplicates}
\end{figure}

\begin{figure}
    \centering
    \includegraphics[width=\linewidth]{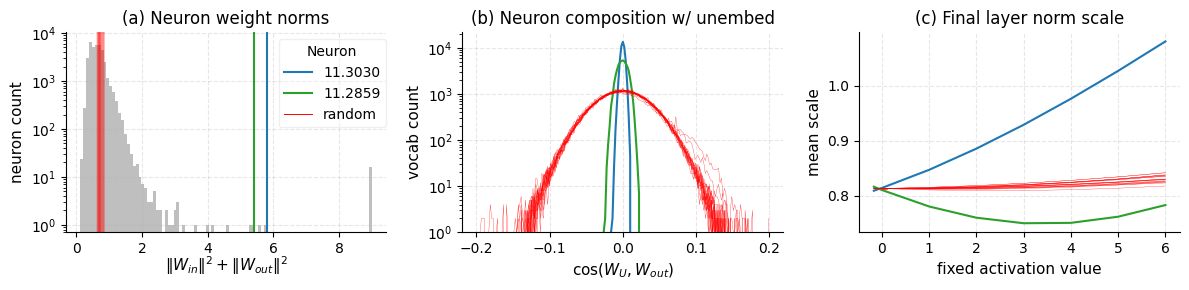}
    \includegraphics[width=\linewidth]{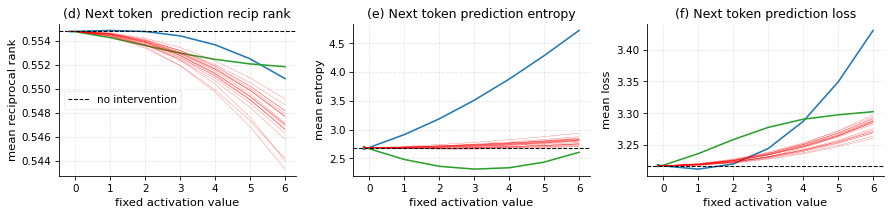}
    \caption{Summary of (anti-)entropy neurons in GPT2-small-a compared to 20 random neurons from final two layers. Entropy neurons have high weight norm (a) with output weights mostly orthogonal to the unembedding matrix (b). When activated, this causes the final layer norm scale to increase dramatically (c) while leaving the relative ordering over the next token prediction mostly unchanged (d). Increased layer norm scale squeezes the logit distribution, causing a large increase in the prediction entropy (e; or decrease for anti-entropy neuron) and an increase or decrease in the loss depending on the model's baseline level of under- or over-confidence (f). Legend applies to all subplots.}
    \label{fig:entropy_interventions_gpt2_small}
\end{figure}

\begin{figure}
    \centering
    \includegraphics[width=\linewidth]{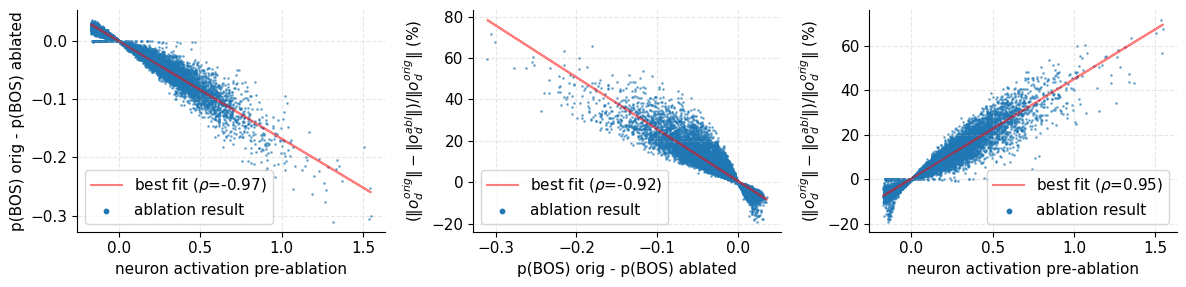}
    \includegraphics[width=\linewidth]{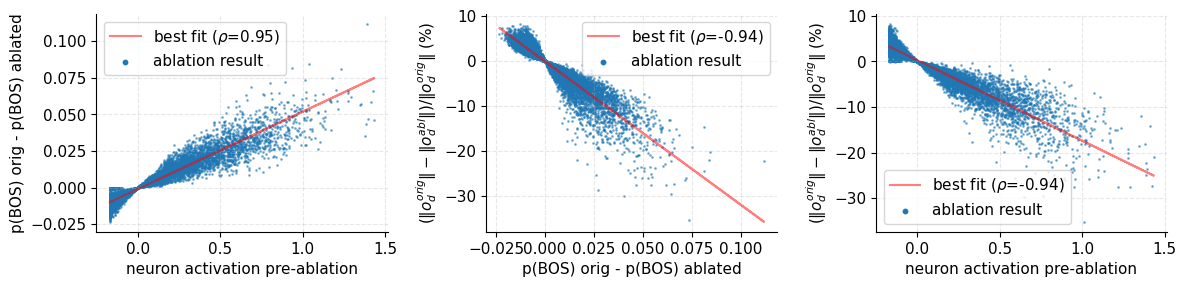}
    \includegraphics[width=\linewidth]{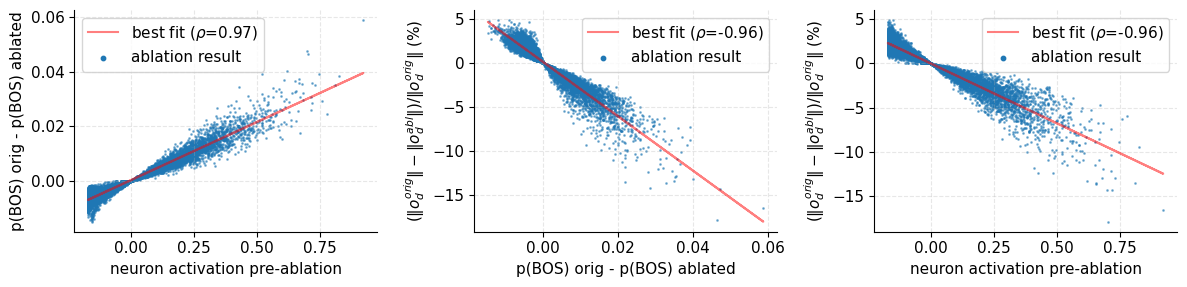}
    \includegraphics[width=\linewidth]{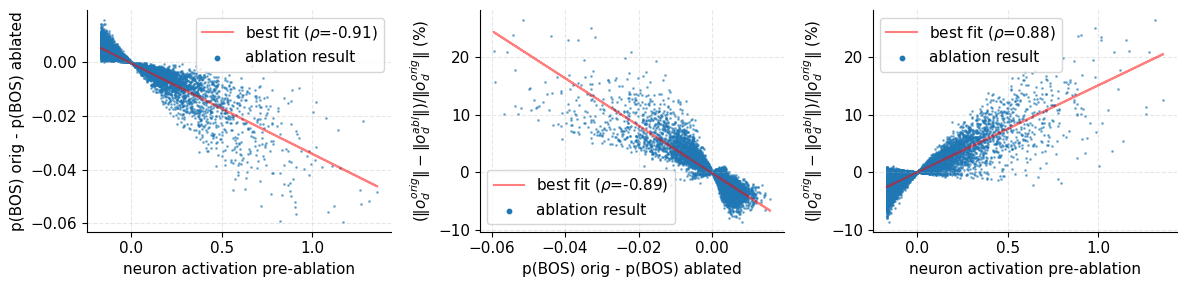}
    \includegraphics[width=\linewidth]{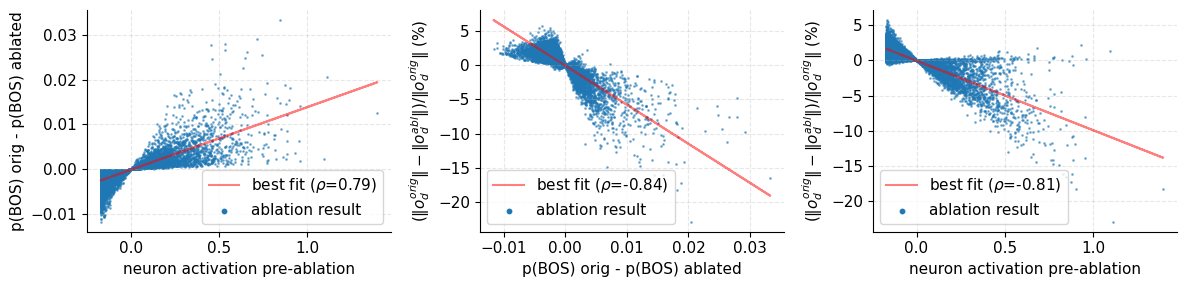}
    
    \caption{Further examples of attention activation and deactivation neurons. Row 1: A15H8 with L14N411, Row 2: A15H8 with L14N2335, Row 3: A15H8 with L14N1625, Row 4: A20H4 with L19N2509, Row 5: A22H7 with L20N2114}
    \label{fig:more_attention_activation_gpt2_medium}
\end{figure}

\end{document}